\documentclass[journal]{IEEEtran}
\usepackage{graphicx}
\usepackage{amsfonts}
\usepackage{color}
\usepackage{bm}
\usepackage[linkcolor=blue,bookmarks=true]{hyperref}
\usepackage{amsmath}
\usepackage{amssymb}
\usepackage{multirow}
\usepackage{makecell}
\usepackage{enumitem}
\usepackage{tabularx}
\usepackage{fixltx2e} 
\usepackage{booktabs} 
\usepackage{xcolor}
\usepackage{url}
\usepackage{graphicx}
\usepackage{amsfonts}
\usepackage{color}
\usepackage{bm}

\ifCLASSINFOpdf
\else
\fi

\begin{document}

\title{Learning Spatio-Appearance Memory Network for High-Performance Visual Tracking}

\author{Fei Xie,
	Wankou Yang$^\star$,
	Kaihua Zhang,
	Bo Liu,
    Wanli Xue,
	Wangmeng Zuo
    
\thanks{$^\star$Corresponding author. Fei Xie and Wankou Yang are with the School of Automation, Southeast University, Nanjin, 210096, China, Email: jaffe03@seu.edu.cn, wkyang@seu.edu.cn.}
\thanks{Kaihua Zhang is with Jiangsu Key Laboratory of Big Data Analysis Technology (B-DAT), Nanjing University of Information Science and Technology, Nanjing, China. Email: zhkhua@gmail.com.}
\thanks{Bo Liu is with JD Finance America Corporation, Mountain View, CA, USA, 94089}%
\thanks{Wanli Xue is with School of Computer Science and Engineering, Tianjin University of Technology, Tianjin, China. E-mail: xuewanli@email.tjut.edu.cn.}%
\thanks{Wangmeng Zuo is with the School of Computer Science and Technology,
	Harbin Institute of Technology, Harbin, 150001, China and is also with
	Peng Cheng Lab, Shenzhen, China, e-mail: (cswmzuo@gmail.com).}%
\thanks{Manuscript received xxx; revised xxx.}}

\markboth{Journal of \LaTeX\ Class Files,~Vol.~14, No.~8, August~2015}%
{Shell \MakeLowercase{\textit{et al.}}: Bare Demo of IEEEtran.cls for IEEE Journals}

\maketitle

\begin{abstract}
Existing visual object tracking usually learns a bounding-box based template to match the targets across frames, which cannot accurately learn a pixel-wise representation, thereby being limited in handling severe appearance variations caused by large-scale deformation, severe scale variation, and heavy occlusion, etc.
To address these issues, much effort has been made on segmentation-based tracking, which learns a pixel-wise object-aware template and can achieve higher accuracy than bounding-box template based tracking.
However, existing segmentation-based trackers are ineffective in learning the spatio-temporal correspondence across frames due to no use of the rich temporal information.
To overcome this issue, this paper presents a novel segmentation-based tracking architecture, which is equipped with a spatio-appearance memory network to learn accurate spatio-temporal correspondence.
Among it, an appearance memory network explores spatio-temporal non-local similarity to learn the dense correspondence between the segmentation mask and the current frame, which can effectively model long-range dependency to capture stable appearance information. Meanwhile, a spatial memory network is modeled as discriminative correlation filter to learn the mapping between feature map and spatial map. The appearance memory network helps to filter
out the noisy samples in the spatial memory network while the latter provides the former with more accurate target geometrical center. This mutual promotion greatly boosts the tracking performance.
Without bells and whistles, our simple-yet-effective tracking architecture sets new state-of-the-arts on the VOT2016, VOT2018, VOT2019, GOT-10K, TrackingNet, and VOT2020 benchmarks, respectively, especially achieving the EAO scores of 0.535 and 0.506 respectively on VOT2016 and VOT2018. Besides, our tracker outperforms the leading segmentation-based trackers  SiamMask and D3S on two video object segmentation benchmarks DAVIS16 and DAVIS17 by a large margin. The source codes can be found at \url{https://github.com/phiphiphi31/DMB}.
\end{abstract}


\IEEEpeerreviewmaketitle

\section{Introduction}

Visual object tracking (VOT) is a fundamental task in computer vision. In general, VOT aims at localizing the target in subsequent frames based the given bounding box in the first frame.
So far, VOT remains challenging due to numerous factors such as deformation, occlusion, and background clutter, to name a few~\cite{otb, kristan2018sixth,lasot}.
Two dominant methodologies of deep learning based VOTs are Siamese correlation networks~\cite{li2018high,siamrpn++,xu2020siamfc++,siamcar,siamban} and discriminative correlation filters (DCFs)~\cite{atom,dimp}, which mainly adopt a bounding box-level target representation, making them limited in exploiting the fine-grained representation of the target that is essential to achieve a high tracking accuracy.
Moreover, the bounding-box representation is appropriate for axis-aligned transformation, but insufficient in handling complex and non-rigid transformation.
To address these issues, estimating pixel-wise target representation is a prerequisite, but it requires segmentation datasets~\cite{YTBVOS} for training which are far less than the tracking datasets, such as TrackingNet~\cite{Trackingnet}, LaSOT~\cite{lasot} and GOT-10K~\cite{got}, due to the extremely laborious annotations.

\begin{figure}[t]
	\centering{\includegraphics[scale = 0.35]{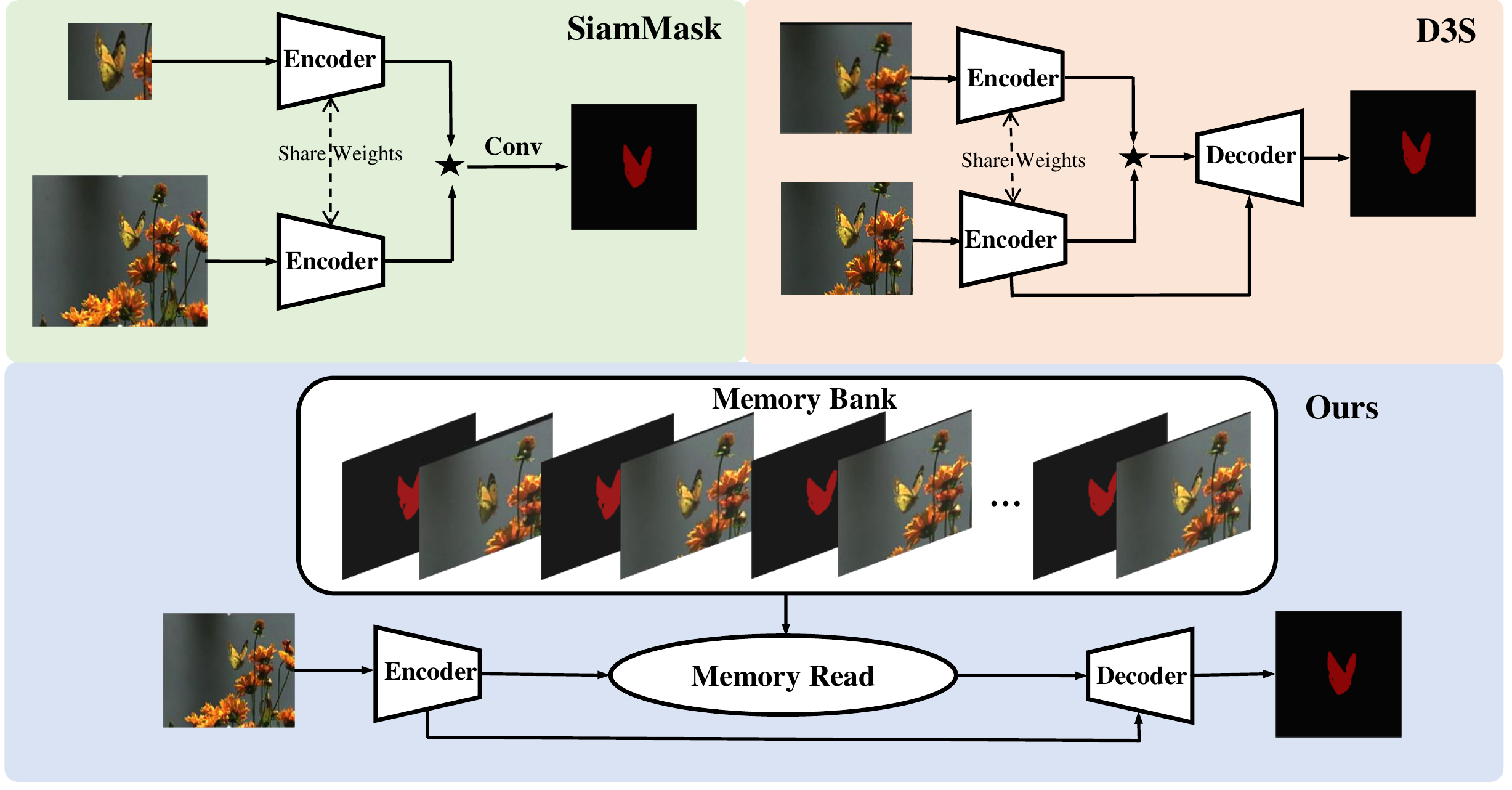}}
	\caption{Comparison of the proposed pipeline with SiamMask and D3S that do not take into account the rich temporal information for matching.}
	\label{fig:novel}
\end{figure}

Several attempts have been made to develop segmentation-based trackers which leverage pixel-wise object-aware representations for matching.
\begin{figure}[t]
	\centering{\includegraphics[scale = 0.40]{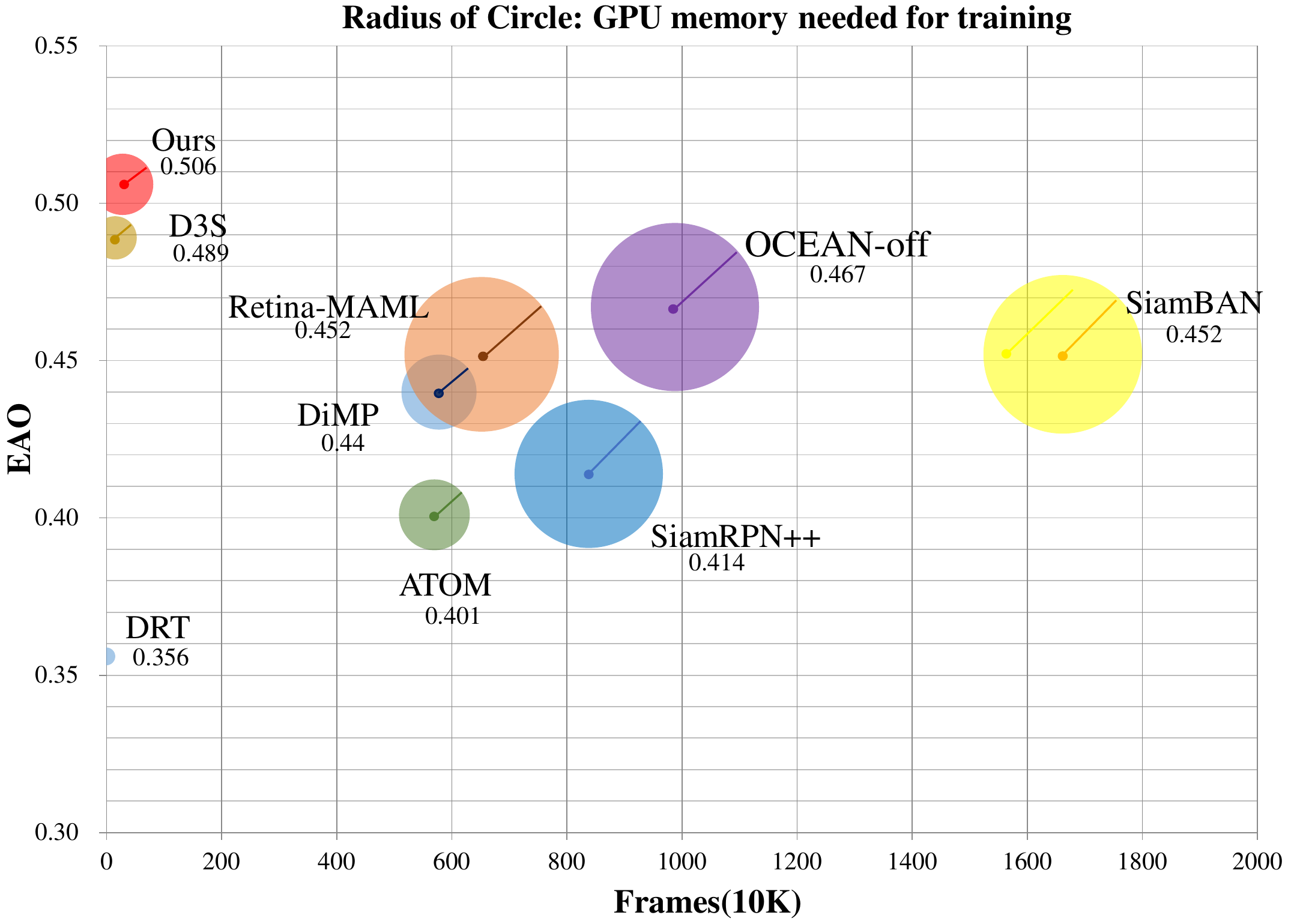}}
	\caption{Comparison of tracking performance and offline training cost with state-of-the-art (SOTA) trackers on VOT2018~\cite{kristan2018sixth}. We visualize the Expected Average Overlap (EAO) with respect to the amount of training frames. The radius of circle denotes the GPU memory needed for training (16GB is needed for our tracker). DRT~\cite{sun2018correlation} is a fully online tracker that achieves the best efficiency but a much lower EAO than ours.}
	\label{fig:babble}
\end{figure}
SiamMask~\cite{siammask} adds a segmentation branch to Siamese architecture that allows for joint learning of bounding box regression and object segmentation in training, but fails to seamlessly integrate them in the tracking stage.
Later, D3S~\cite{lukezic2020d3s} introduces a segmentation branch following VideoMatch~\cite{hu2018videomatch} and further combines online DCF~\cite{bolme2010visual} to fuse target classification and pixel-wise segmentation results during inference.
While the DCF can be updated to cope with appearance variations across frames, little study has been given to consider temporal information in the segmentation branch during tracking.
Moreover, the segmentation branch in SiamMask is offline trained, while that in D3S is initialized using the first frame and then fixed during inference.
Thus, both SiamMask and D3S fail in utilizing the useful temporal information to enhance the segmentation branch (see Fig.~\ref{fig:novel}),  and may lead to model under-fitting during tracking.
%

To overcome the limitations of SiamMask and D3S, this paper exploits spatio-appearance memory networks to capture long-range spatio-temporal information to learn dense correspondence for segmentation based tracking.
%
%
%
Specifically, we design an appearance memory network (AMN) to adapt the segmentation branch to temporal appearance variation while avoiding model drifting.
In the case of template-based or one-shot detection based trackers,
their memory encoded by target template feature map is typically too small, and is not compartmentalized enough
to accurately remember facts from the past during handling long video tasks.
To address this issue, we store keys and values of continuous frames in the AMNs, and design a memory reader to compute the spatio-temporal attention to previous frames for each pixel in the query image (i.e., the current frame).
Thus, albeit the network parameters of the memory module are fixed, we can dynamically update the memory network to achieve better tradeoff between model generalization and flexibility (see Fig.~\ref{fig:novel}).
%
%
To encode precision spatial information for accurate localization, we further construct a spatial memory network (SMN) by using DCF to model the mapping between feature map and spatial map.
The SMN helps filter out the noisy samples in the AMN while the AMN provides the SMN with more accurate target geometrical center. This mutual promotion greatly boosts the tracking performance.
Besides, we leverage a box-to-segmentation training and testing strategy to mitigate inaccurate representation of bounding box initialization during tracking.
Extensive experiments show that our tracker sets a new state-of-the-art on the popular tracking benchmarks including VOT2016, VOT2018, VOT2019, VOT2020, GOT-10K, and TrackingNet.
%
%
Moreover, for the video object segmentation (VOS) task, our tracker also surpasses the leading discriminative segmentation-based trackers D3S and SiamMask by a large margin on the DAVIS16 and DAVIS17 benchmarks.
In comparison to the SOTA template-based trackers, our approach can reduce the training data by more than an order of magnitude with improved tracking performance (see Fig.~\ref{fig:babble}).
%

The main contributions of this work are three-fold:


\begin{itemize}
	\item We present a novel AMN design that can capture the long-term appearance changes to effectively enhance the segmentation branch of the segmentation-based tracking.
	
	\item We design a DCF-based SMN module and collaborate it with the AMN to form our novel spatio-appearance memory network design, which can effectively bridge the gap between VOT and VOS.
	\item Extensive experiments on six challenging tracking benchmarks demonstrate that our tracker achieves SOTA performance. Meanwhile, for the VOS task, ours outperforms D3S and SiamMask on DAVIS16 and DAVIS17 by a large margin.
	
\end{itemize}

\section{Related Work}
\subsection{Bounding-box based VOT}
The state-of-the-art template-based trackers generally can be grouped into two categories, i.e., Siamese correlation networks and DCF-based tracking approaches.
Representative Siamese correlation trackers such as RPN-based~\cite{ren2015faster} trackers~\cite{li2018high,siamrpn++} and anchor-free trackers ~\cite{xu2020siamfc++,siamban} usually consist of a classification branch for foreground-background estimation and a regression branch for box refinement.
%
%
Recent filter-based trackers combine DCF~\cite{bolme2010visual} with modified IoU-Net~\cite{jiang2018acquisition}.
For example, ATOM~\cite{atom} and DiMP~\cite{dimp} utilize DCF for coarse localization, and exploit IoU prediction network for bounding box refinement.
However, template-based trackers generally are training data-hungry to generalize well to previously unseen categories.
Moreover, either the  regression  branch in Siamese trackers or the IoU prediction network in ATOM are offline trained and cannot adapt to temporal appearance changes during tracking.
Furthermore, the bounding box representation of target restricts their ability in handling complex  non-rigid transformation.
In comparison, our approach uses the discriminative segmentation tracking framework to handle complex transformation and alleviate the data-hungry issue, and incorporate spatio-appearance memory network to cope with temporal appearance changes of target.

\begin{figure*}[t]
	\centering{\includegraphics[scale = 0.53]{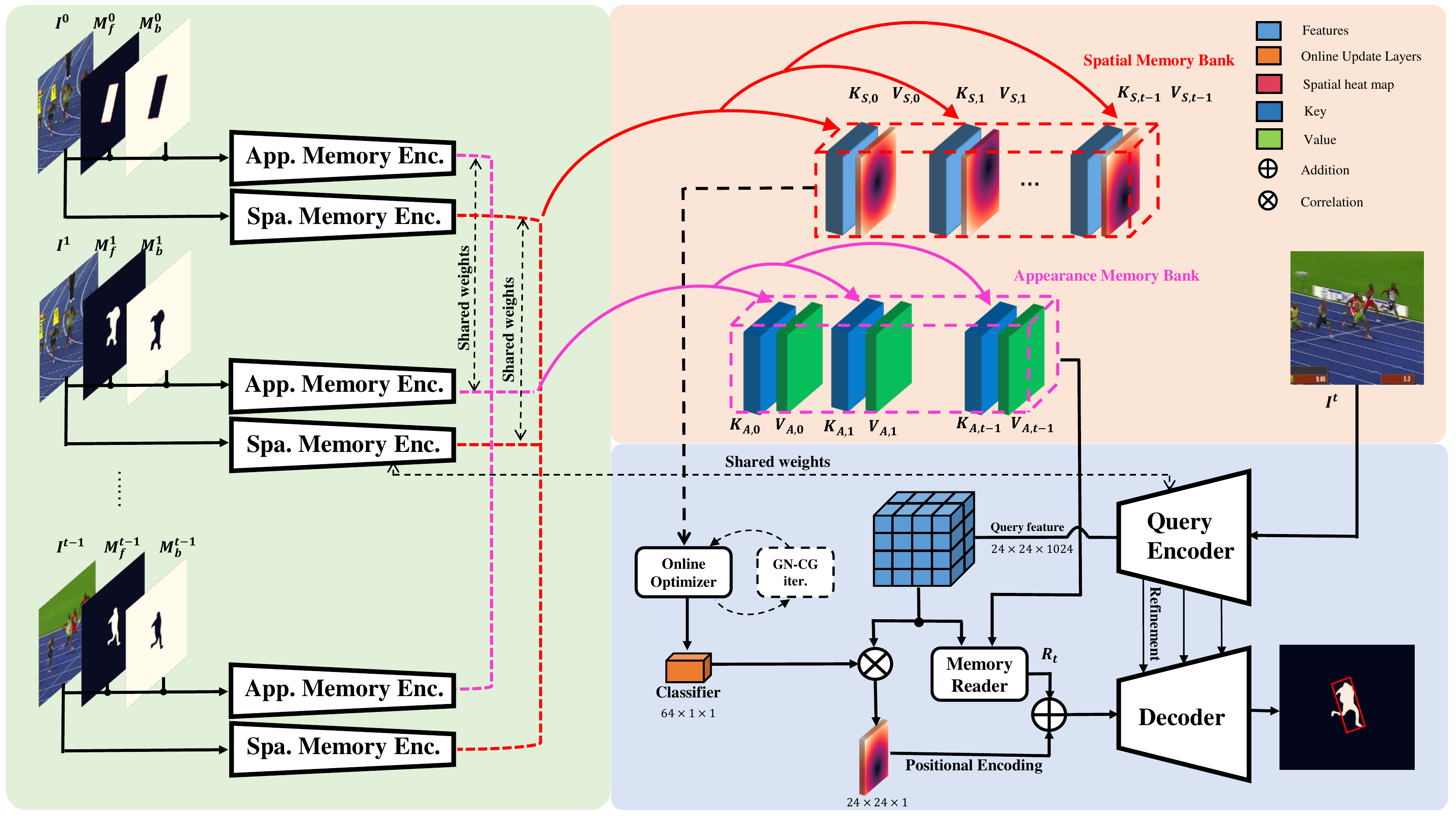}}
	\caption{Overview of our spatio-appearance memory network. The network consists of two memory networks to handle temporal target appearance changes. One is the AMN which makes dense non-local matching in temporal domain to capture stable appearance information. The other is the SMN which trains two convolutional layers by Gauss-Newton (GN) and Conjugate Gradient (CG) iterations. The decoder accepts the fused read-out features from the spatio-appearance memory networks to output the segmentation mask. Finally, the bounding box of the tracked target is estimated from the segmentation mask.}
	\label{fig:pipeline}
\end{figure*}

\subsection{Segmentation based VOT}
VOS methods~\cite{OnAVOS, FAVOS, OSVOS} usually are slow in speed and are ineffective in handling the challenging factors in tracking scenarios, e.g., distractors and fast motion.
It is mainly because the VOS task considers segmentation of large objects with limited appearance changes in short videos. SiamMask~\cite{siammask} attempts to unify tracking and segmentation by adding a class-agnostic segmentation branch to detection-based tracker.
SiamRCNN~\cite{siamrcnn} uses an well-trained segmentation model to estimate the mask in the box which considers the predicted bounding box as hard spatial constraints. Similarly, many VOT methods such as OceanPlus~\cite{Kristan2020a} and SiamMargin~\cite{Kristan2020a} add extra segmentation model after predicting bounding box to improve tracking accuracy.
D3S~\cite{lukezic2020d3s} uses the DCF as the classification branch and a geometrically invariant template-based model for object segmentation.
Our method also adopts the segmentation tracking architecture, and the dual memory networks are introduced to utilize temporal information.

\subsection{Memory Network for Video Analysis}
Memory network is a kind of neural network that has external memory where information can be stored and read by purpose. Memory network is widely used to solve simple logical reasoning problem in natural language processing such as NTM (Neural Turing Machine)~\cite{ntm} and MemNN (Memory Neural Networks)~\cite{memNN}.
Recently, memory network has exhibited its merits in temporal modeling for video tasks.
In visual tracking, MemTrack~\cite{yang2018learning} uses a dynamic memory network to adapt the template to appearance variations.
STM~\cite{oh2019video} applies memory networks to semi-supervised VOS and achieves appealing performance.
MAST~\cite{MAST} exploits the memory module as spatio-temporal non-local attention in self-supervised VOS.
However, most of those methods apply in VOS task typically consider large targets with low background distractor presence. However, in object tracking benchmarks, the size of targets is relatively small and the motion is relatively large. The VOS methods thus work poorly in challenging tracking scenarios.
In this work, we present a spatio-appearance memory network to exploit the temporal appearance changes that is able to bridge the gap between VOT and VOS.

\section{Proposed Approach}
%
%
In the following, we first introduce the overall pipeline of our approach in~Section~\ref{sec:pipepline}, and then explain the appearance memory network (Section~\ref{sec:Appearance Memory Network}), spatial memory network (Section~\ref{sec:Saptial Memory Network}), and decoder (Section~\ref{sec:decoder}) in details. The interactions between the two networks are explained in Section~\ref{sec:benefit}, and finally, we provide the detailed design of the loss function in Section~\ref{sec:loss}.
%


\begin{figure*}[t]
	\centering{\includegraphics[scale = 0.50]{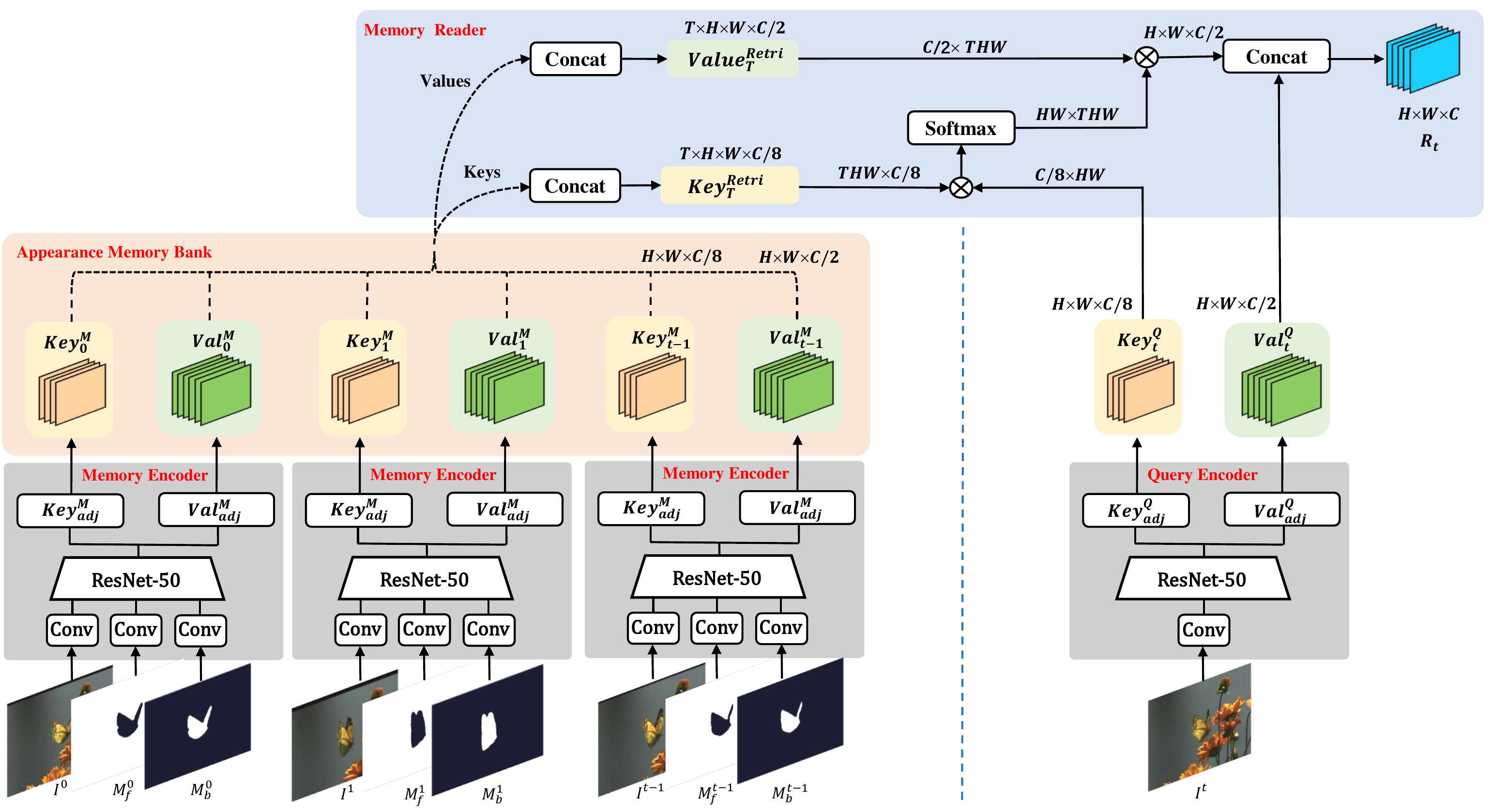}}
	\caption{Overview of the AMN. Each continuing frame and its foreground-background mask generates corresponding key and value through appearance memory encoder. Query frame $I_t$ will be encoded into query $Q_{t}$ and value $V_{A,t}$ embedding. A dense non-local matching operation will be performed between query $Q_{t}$ and stored memory keys  $\{K_{A,0}, ..., K_{A,t-1}\}$. The retrieved value $V_{Q,t}$ from read operation will be concatenated with query value $V_{A,t}$ as the read-out value $R_{t}$. Then, the read-out value $R_{t}$ will be fed into decoder for final mask prediction. }
	\label{fig:keyval}
\end{figure*}

\subsection{Overall Pipeline}
\label{sec:pipepline}

Fig.~\ref{fig:pipeline} illustrates the pipeline of our approach.
Each frame $I_{t}$ is embedded into two triplets $\left(Q_{t}, K_{A,t}, V_{A,t}\right)$ and $\left(Q_{t}, K_{S,t}, V_{S,t}\right)$.
As in memory network~\cite{weston2015memory}, $Q$, $K$, and $V$ refer to Query, Key, and Value, respectively.
For the tracking and segmentation of current frame $I_t$, an appearance memory encoder $ {\bf Enc}_M^{A}$ is used to
compute the appearance memory key and value representation pairs $\{(K_{A,0}, V_{A,0}), ..., (K_{A,t-1}, V_{A,t-1}) \}$ for the previous frames $\{I_0, ..., I_{t-1}\}$.
Meanwhile, a spatial memory encoder ${\bf Enc}_M^{S}$ is introduced to extract the spatial memory keys $\{K_{S,0}, ..., K_{S,t-1} \}$.
Following conventional tracking setting, the values $\{V_{S,0}, ..., V_{S,t-1} \}$ of the spatial memory are computed based on the annotation of the first frame and the predicted target bounding boxes of the previous frames.
Moreover, a query encoder ${\bf Enc}_Q$ is designed to obtain the query $Q_t$ and the query value $V_{Q,t}$ for the current frame $I_t$.
Furthermore, the appearance memory reader module is adopted to generate the value $V_{A,t}$ for the current frame.
As for spatial memory, we take DCF as a memory module, and use it to generate target location map.
Subsequently, $V_{A,t}$, $V_{Q,t}$, the target location map, and the query encoder features are incorporated into a decoder to predict the segmentation mask of $I_t$.
Finally, the target bounding box can be estimated from the segmentation mask. To adapt target appearance variations over time during tracking, the memory keys and values are updated online and added to the appearance and spatial memory networks.

\subsection{Appearance Memory Network}
\label{sec:Appearance Memory Network}
Fig.~\ref{fig:keyval} shows the architecture of our appearance memory network that includes AMN and memory reader. Analogous to conventional memory network~\cite{memNN, STM}, our memory network consists of memory encoder ${ \bf Enc}_M^{A}$, query encoder ${\bf Enc}_Q$, and memory reader.
In particular, for each of the previous frames, the memory encoder takes the image $I$ and the foreground as well as the background segmentation masks $\{M_f, M_b\}$ as the input to produce the key and the value.
And the current frame $I_t$ is fed into query encoder to obtain query $Q_{t}$ and query value $V_{Q,t}$.
Then, query $Q_{t}$ is passed into the memory reader to obtain the retrieved value $V_{A,t}$ from AMN.
Finally, $V_{A,t}$ and $V_{Q,t}$ are concatenated to form the read-out value $R_t$.
Next, we introduce the memory encoder, query encoder and memory reader in detail.


\textbf{Memory Encoder.}
The input of memory encoder involves three components, i.e., an RGB frame, the foreground and background segmentation masks with probability between 0 and 1.
Each component first goes through three convolutional layers individually and then be summed and fed into the backbone.
Here we take ResNet-50~\cite{he2016deep} as the backbone for both the memory encoder and the query encoder, and use the Conv4\_e layer as the common feature map $f_M$ for computing the key and value.
Then, the key and value can be obtained by respectively deploying their own convolutional layer on the commen feature map $f_M$
\begin{equation}K_{A}={\bf Key}_{A}\left(f_{M}\right), \quad V_{A} = {\bf Val}_{A}\left(f_{M}\right).\end{equation}
During tracking, keys and values from all previous frames are stacked along the temporal order and are stored in the AMN.

\textbf{Query Encoder.}
The query encoder ${\bf Enc}_Q$ takes the current frame $I_t$ as the input to produce the query $Q_t$ as well as the query value $V_{Q,t}$.
Analogous to memory  encoder, we use the Conv4\_e layer of  ResNet-50 as the common feature map $f_Q$.
To generate the query $Q_t$, a convolutional block is applied to reduce the number of channels to the $1/8$ of $f_Q$.
The channel number of query value $V_{Q,t}$ is a half of $f_Q$

\begin{equation} Q_t= {\bf Que}_{A}(f_Q),  V_{Q,t} = {\bf Val}_{Q} (f_Q). \end{equation}

\textbf{Memory Reader. }
In the memory reader module, the keys and values $\{(K_{A,0}, V_{A,0}), ..., (K_{A,t-1}, V_{A,t-1}) \}$ of all previous frames, and the query and query value $(Q_t, V_{Q,t})$ of the current frame are used to produce the read-out value $R_t$. In particular, the similarities between query $Q_t$ and keys $\{K_{A,0}, ..., K_{A,t-1})\}$ are utilized to measure the spatial and temporal non-local correspondence, which is then used to generate the retrieved value $V_{A,t}$ for capturing temporal appearance changes.
Then, the retrieved value $V_{A,t}$ is computed based on the non-local attention mechanism formulated as follows
\begin{equation}{V}_{A,t}^{i} = \sum_{j} \sum_{k=1}^{t-1} A_{t}^{i,j,k} V_{A,k}^{j},  \end{equation}

\begin{equation}A_{t}^{i,j,k}=\frac{\exp \left\langle Q_{t}^{i}, K_{A, k}^{j}\right\rangle}{\sum_{p} \sum_{k=1}^{t-1} \exp \left\langle Q_{t}^{i}, K_{t-1}^{p}\right\rangle},
	\label{eq2}\end{equation}
where $i$, $j$, and $p$ denote the spatial position of feature map, $k$ denotes the index of a frame, and $\langle\cdot, \cdot\rangle$ denotes the dot product between two vectors. %
We note that $A_{t}^{i,j,k}$ measures the similarity between query and keys in spatio-temporal domain.
Thus, albeit the  network  parameters  of  the  appearance memory  module  are fixed, the memory network can be updated during tracking, and the non-local spatio-temporal matching further makes it feasible for coping with temporal appearance variation of target.
Furthermore, for enhancing the retrieved value $V_{A,t}$, we concatenate it with the query value $V_{Q,t}$ to obtain the read-out value
\begin{equation}R_{t}=concat \left[V_{A,t}, V_{Q,t}\right],\end{equation} where $concat[\cdot, \cdot]$ denotes the concatenation operation.


In contrast to MAST~\cite{MAST} where the RGB image or segmentation mask are adopted as the value, we predict the appearance value as embedding feature map.
And we encode both the query and the query value, and the later is further concatenated with the retrieved value to get the read-out value.

\begin{figure}[t]
	\centering{\includegraphics[scale = 0.38]{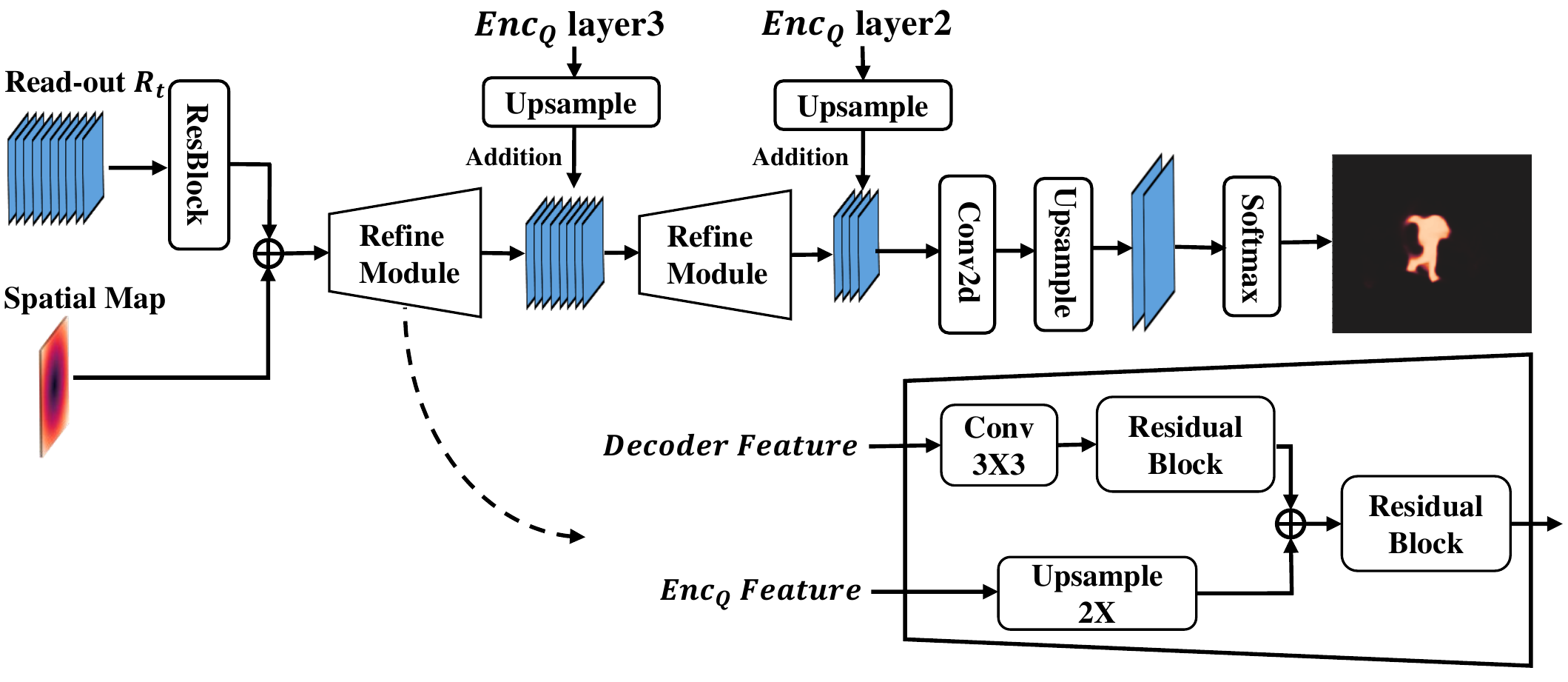}}
	\caption{Architecture of the decoder.}
	\label{fig:decoder}
\end{figure}

\subsection{Spatial Memory Network}
\label{sec:Saptial Memory Network}
Inspired by~\cite{atom}, we introduce the SMN to capture the temporal appearance changes of target to improve localization performance.
In particular, the query encoder ${\bf Enc}_Q$ in the AMN shares weights with the memory encoder and the query encoder for SMN.
Let $x_k = {\bf Enc}_Q(I_k)$ be the feature map for a previous frame $I_k$, and $y_k$ be the corresponding spatial label.
The DCF model can then be formulated as
\begin{equation} { f^* = \arg \min_{f} } \sum_{k=0}^{t-1} \sum_{p} \left\| \langle x_k^p , f \rangle -y_{k}\right\|_{2}^{2} +\lambda \|f\|_{2}^{2} . \end{equation}
With the kernel tricks, we have
\begin{equation} R_{S,t}^i = \langle f^*, x_t^i \rangle = \sum_{k=0}^{t-1} \sum_{j} V_{S,k}^j A_{S,t}^{i,j,k}, \end{equation}

\begin{equation}A_{S, t}^{i,j,k}= {\left\langle x_{t}^{i}, x_{k}^{j}\right\rangle},
\end{equation}
where $i$, $j$, and $p$ denote the spatial position.
We note that $\{x_k| k = 0, ..., t-1 \}$ and $x_t$ can be explained as the keys and query, while $V_{S,k}$ and $R_{S,t}$ are the value and read-out value in SMN.
For model update, we employ the fast Gauss-Newton (GN) and Conjugate Gradient (CG) algorithm to train $A_{S, t}^{i,j,k}$ online during inference.
Thus, DCF can be explained as a special implementation of memory module to store the mapping between feature map and the read-out spatial map $R_{S,t}$.
Moreover, the spatial map $R_{S,t}$ can serve as the spatial encoding of target localization, which is complementary to the read-out value in AMN.
In our approach, we combine AMN and SMN to constitute the spatio-appearance memory network for improving segmentation and tracking performance.

\subsection{Decoder}
\label{sec:decoder}
Fig.~\ref{fig:decoder} illustrates the network architecture of the decoder.
The read-out values of spatio-appearance memory network are feature map and spatial map, which are then fed into the decoder module for predicting the final mask of the target. The spatial map serves as positional encoding for the feature map from AMN. The full architecture of decoder can be referred to \cite{STM} and \cite{frtm}. The refinement module contains two residual blocks to fuse the decoder feature with template feature. In this stage, the decoder feature with spatial attention is more discriminative towards challenging scenarios in tracking. 
%
%
%

\begin{figure}[t]
	\centering{\includegraphics[scale = 0.30]{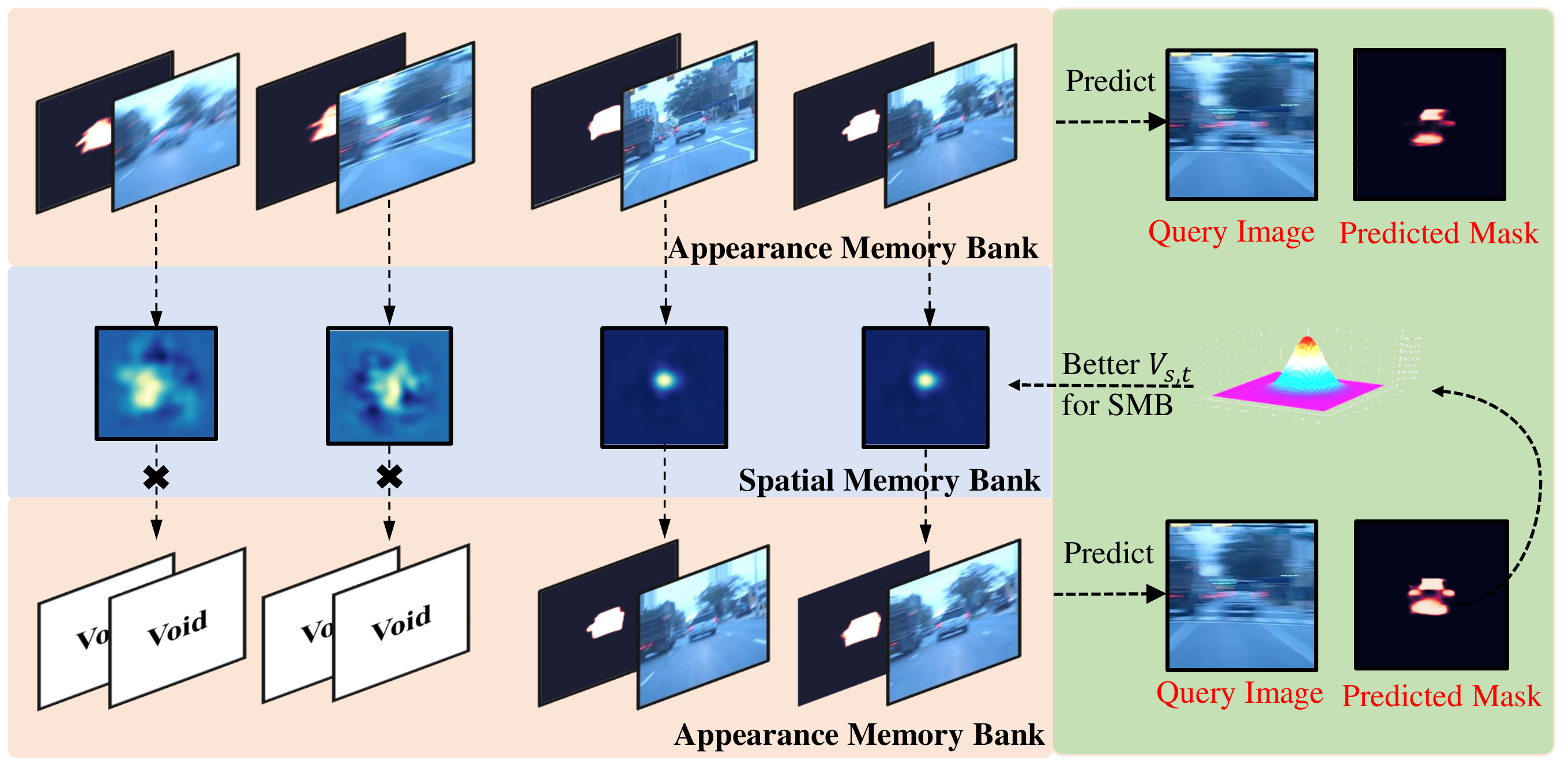}}
	\caption{Visualization on the benefit from filtered samples.}
	\label{fig:filter}
\end{figure}

\subsection{Interactions between AMN and SMN}
\label{sec:benefit}
In general, the SMN is complementary to the AMN and can collaborate to improve segmentation and tracking performance. We also present an elaborate design to make the two networks cooperate well. Due to extreme challenging factors like blur and occlusion, we evaluate the state of the spatial map from the SMN to filter the noisy samples out and store high quality samples in the AMN. We construct a queue which has fixed maximum length $L$ for storing uncertainty values to evaluate the quality of current frame. The uncertainty value of each frame $\{U_{0}, U_{1},..., U_{L-1}\}$ is to invert the peak value of the spatial map from the SMN. Then, the decision to segment the current query frame $I_{t}$ will be determined by the comparsions of current uncertainty value $U_{t}$ and the average uncertainty value of stored queue $U_{avg, t}$. In the case that current uncertainty value $U_{t}$ is larger than the threshold $Threshold =10$, the key and value current frame will be removed directly. These processes are formulated as

\begin{equation}
U_{t}=\frac{1}{\operatorname{Max}\left(V_{S, t}\right)},
\end{equation}

\begin{equation}
U_{avg, t}=\frac{1}{L}\sum_{k=1}^{L-1} U_{k},
\end{equation}

\begin{equation}
D_{t}=\left\{\begin{array}{cc}
\text {preserved} &  \text {if } U_{t} < U_{avg, t} \\
\text {removed} & \text {if } U_{t} > U_{avg, t} \\
\text {removed} & \text {if } U_{t} > Threshold{set}.
\end{array}\right.
\end{equation}

The queue for storing the uncertainty values will be maintained to be fixed maximum length $L$. The earliest uncertainty value will be removed if the queue is full. As shown in Fig.~\ref{fig:filter}, the predicted mask from filtered samples is more accurate. On the contrary, the samples $\{V_{S,k}| k = 0, ..., t-1 \}$ stored in SMN are benifit from the geometric center generated by pixel-level representations. The localization of SMN can be more robust because of the geometical robustness of target center.

\subsection{Loss Function}
\label{sec:loss}

To optimize the proposed framework, instead of exploiting large scale of tracking training sets, such as TrackingNet~\cite{Trackingnet}, LaSOT~\cite{lasot} and GOT-10K~\cite{got}, we only use segmentation training sets which significantly reduces the training cost. In the design of loss, we formulate our training loss as:
\begin{equation}\mathcal{L}=\mathcal{L}_{B C E}\left(y^{l}, y^{p}\right)+\lambda \mathcal{L}_{I o U}\left(y^{l}, y^{p}\right),\label{eqloss}\end{equation}
where $\mathcal{L}_{B C E}$ indicates the binary cross-entropy loss, $\mathcal{L}_{I o U}$ indicates the standard IoU loss. $y^{p}$ in Eq.~\ref{eqloss} is the predicted mask and $y^{l}$  is the mask label.

\begin{figure}[t]
	\centering{\includegraphics[scale = 0.28]{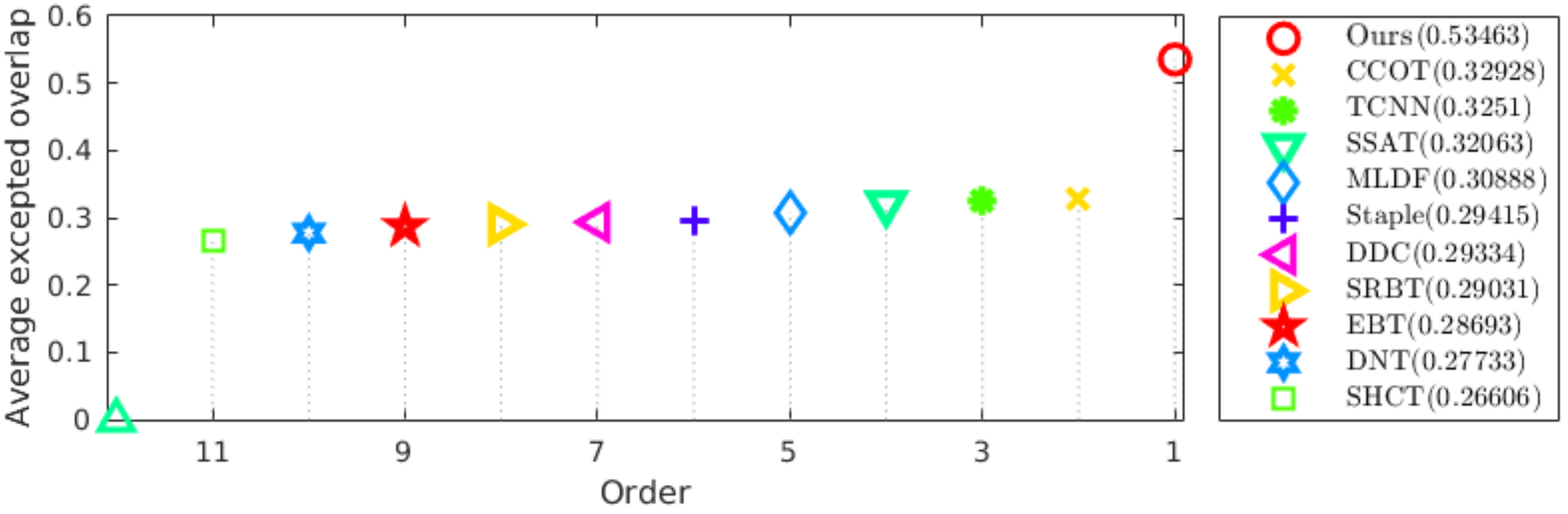}}
	\caption{EAO ranking plot on VOT2016.}
	\label{fig:16eao}
\end{figure}

\begin{figure}[t]
	\centering{\includegraphics[scale = 0.56]{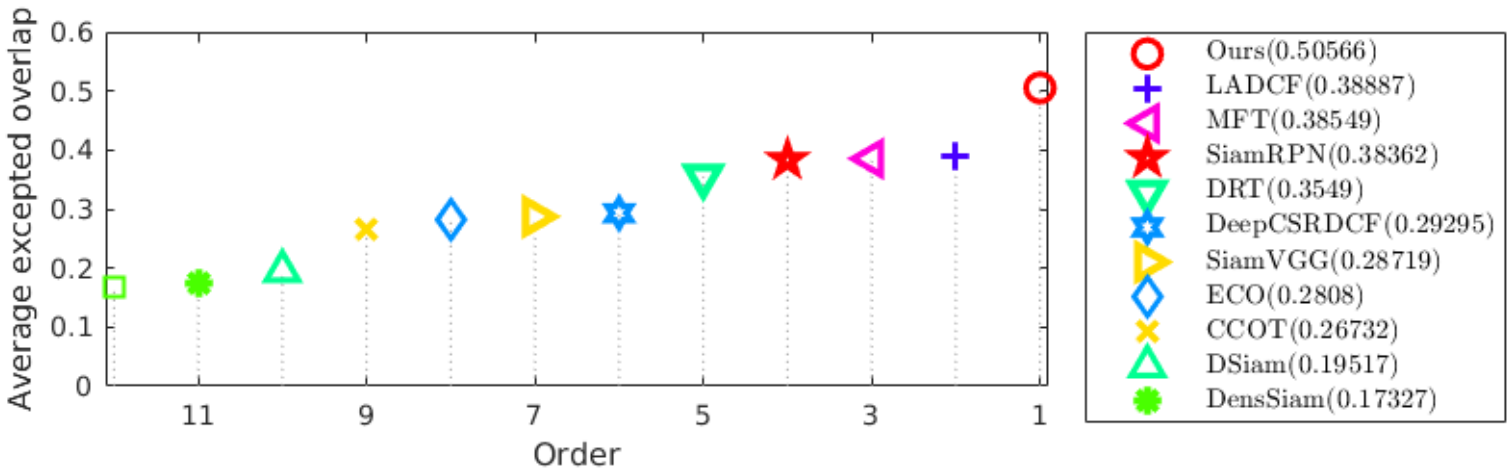}}
	\caption{EAO ranking plot on VOT2018.}
	\label{fig:18eao}
\end{figure}

\begin{figure}[t]
	\centering{\includegraphics[scale = 0.56]{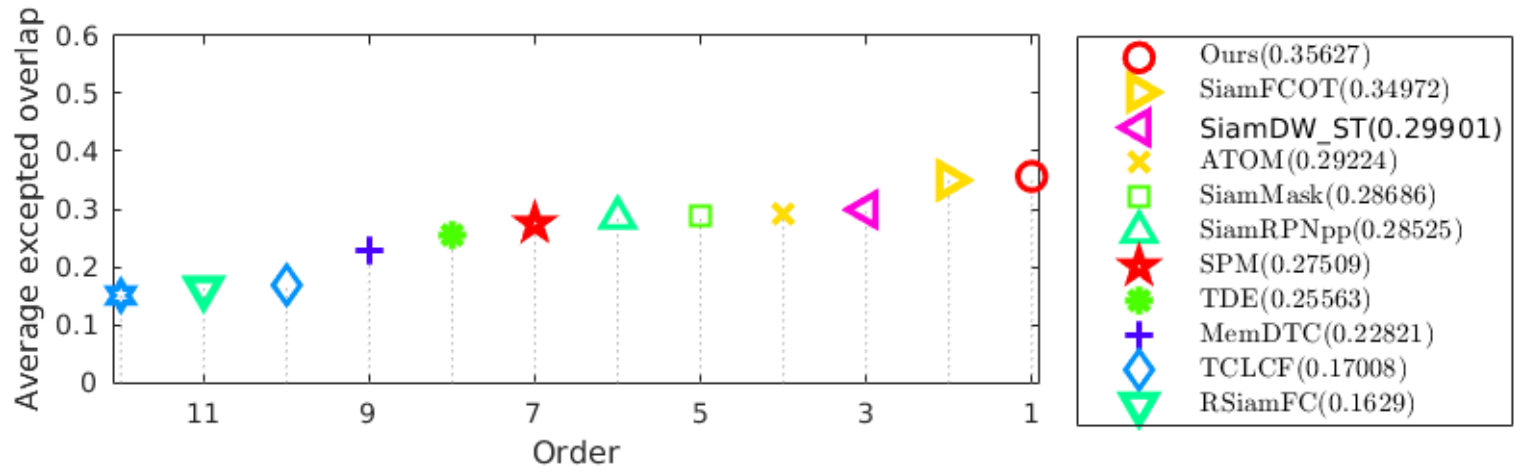}}
	\caption{EAO ranking plot on VOT2019.}
	\label{fig:19eao}
\end{figure}

\section{Experiments}
In this section, we first introduce the implementation details in Section~\ref{sec:Implementation}, and then introduce the evaluation methods and analyze the compared results in details in Section~\ref{sec:evaluation}. Finally, we conduct ablation study to validate the effectiveness of the key design component in Section~\ref{sec:ablation}.
\subsection{Implementation Details} \label{sec:Implementation}
\paragraph{Training Phase}
We firstly use image datasets instead of video sequences for training. We use ResNet-50 as backbone network which is pretrained in ImageNet. Similar to the training process in \cite{perazzi2017learning} and \cite{STM}, we use image datasets with object masks(~\cite{shetty2016application},~\cite{lin2014microsoft},~\cite{shi2015},~\cite{li2014secrets},~\cite{cheng2014global}) to train our network. We apply image augmentations like affine, flip and blur to the same image for generating a sequence of three images.

After finishing training the encoder block, we use Youtube-VOS~\cite{xu2018youtube} and freeze the gradients of the encoder to train the decoder. We randomly pick 3 temporally ordered frames from one video sequence and apply recurrent training strategy. The first frame and its mask are fed into memory encoder. The prediction of second frame is then stored in AMN for predicting the third frame. Then, the loss will be accumulated and backpropogated. We use the cropped 384x384 patch size for training. We minimize the cross-entropy loss and mask IoU loss using Adam optimizer~\cite{kingma2014adam} with a fixed learning rate of $10^{-5}$. The batch size is set to be 4. First-stage training process takes 120 epochs and decoder training takes 40 epochs using four NVIDIA TITAN XP GPUs.
		
\begin{table*} [t]
	\begin{center}
		\tabcolsep=2pt
		\resizebox{2.1\columnwidth}{!}{ %
			\begin{tabular}{c c  c c c c c c c c c c }
				\toprule
				\multicolumn{2}{c}{Trackers} &
				
				\begin{tabular}{c} TCNN~\cite{TCNN} \\  \end{tabular} &
				\begin{tabular}{c} CCOT~\cite{CCOT} \\  \end{tabular} &
				\begin{tabular}{c} UpdateNet~\cite{updatenet} \\  \end{tabular} &
				\begin{tabular}{c} SPM \cite{spm}\\  \end{tabular} &
				\begin{tabular}{c} SiamMask-opt \cite{siammask} \\ \end{tabular}  &
				\begin{tabular}{c} SaimRPN++ \cite{siamrpn++} \\   \end{tabular}  &
				\begin{tabular}{c} ATOM \cite{atom}\\   \end{tabular}  &
				\begin{tabular}{c} D3S \cite{lukezic2020d3s}\\  \end{tabular}  &
				\begin{tabular}{c} Ours \end{tabular} \\
				
				\midrule
				\multirow{3}{*}{VOT-16}
				& Acc.$\uparrow\ $
				& 0.55 & 0.54
				& 0.61 & 0.62
				& {\color{blue}0.67} & 0.64
				& 0.61 & {\color{green}0.66}
				&{\color{red}0.684} \\
				
				& Rob.$\downarrow\ $
				& 0.268 & 0.238
				& 0.21 & 0.21
				& 0.23 & 0.20
				& {\color{green}0.18} & {\color{blue}0.131}
				& {\color{red}0.121}  \\
				
				& EAO$\uparrow\ $
				& 0.325 & 0.331
				& {\color{green}0.481} & 0.434
				& 0.442 & 0.464
				& 0.430 & {\color{blue}0.493}
				& {\color{red}0.535} \\
				
				\bottomrule
		\end{tabular}}%
	\end{center}
	\caption{Results on VOT2016. Top-3 results in each row are colored in red, blue and green, respectively.}
	\label{tab:vot16_results}
\end{table*}

\begin{table*} [!h]
	\begin{center}
		\tabcolsep=2pt
		\resizebox{2.1\columnwidth}{!}{ %
			\begin{tabular}{c c  c c c c c c c c c c }
				\toprule
				\multicolumn{2}{c}{Trackers} &
				
				\begin{tabular}{c} SiamMask\_opt~\cite{siammask} \\  \end{tabular} &
				\begin{tabular}{c} SiamRPN++~\cite{siamrpn++} \\  \end{tabular} &
				\begin{tabular}{c} ATOM~\cite{atom} \\  \end{tabular} &
				\begin{tabular}{c} DiMP-50~\cite{dimp} \\ \end{tabular}  &
				\begin{tabular}{c} SiamBAN~\cite{siamban} \\   \end{tabular}  &
				\begin{tabular}{c} D3S~\cite{lukezic2020d3s} \\   \end{tabular}  &
				\begin{tabular}{c} Ocean-off~\cite{ocean} \\ \end{tabular}  &
				\begin{tabular}{c} DCFST~\cite{DCFST} \\  \end{tabular}  &
				\begin{tabular}{c} Ours \end{tabular} \\
				
				\midrule
				\multirow{3}{*}{VOT-18}
				& Acc.$\uparrow\ $
				& {\color{blue}0.642} & 0.604
				& 0.590 & 0.597
				& 0.597 & {\color{green}0.64}
				& 0.598  & -
				&{\color{red}0.652} \\
				
				& Rob.$\downarrow\ $
				& 0.295 & 0.234
				& 0.203 & {\color{green}0.152}
				& 0.178 & {\color{blue}0.150}
				& 0.169 & -
				& {\color{red}0.145}  \\
				
				& EAO$\uparrow\ $
				& 0.387 & 0.417
				& 0.401 & 0.440
				& 0.452 &{\color{blue}0.489}
				& {\color{green}0.467} & 0.452
				& {\color{red}0.506} \\
				
				\bottomrule
		\end{tabular}}%
	\end{center}
	\caption{Results on VOT2018. Top-3 results in each row are colored in red, blue and green, respectively.}
	\label{tab:vot18_results}
\end{table*}

\begin{table*} [!h]
	\begin{center}
		\tabcolsep=2pt
		\resizebox{2.1\columnwidth}{!}{ %
			\begin{tabular}{c c  c c c c c c c c c c }
				\toprule
				\multicolumn{2}{c}{Trackers} &
				\begin{tabular}{c} SiamDW\_ST \cite{siamdw} \\  \end{tabular} &
				\begin{tabular}{c} SiamMask ~\cite{siammask}\\  \end{tabular} &
				\begin{tabular}{c} SiamRPN++ \\  \end{tabular} &
				\begin{tabular}{c} ATOM \\  \end{tabular} &
				\begin{tabular}{c} Retina-MAML \cite{Wang2020TrackingBI}  \\ \end{tabular}  &
				\begin{tabular}{c} SiamFCOT \cite{Kristan2019a} \\   \end{tabular}  &
				\begin{tabular}{c} Ocean-off \\ \end{tabular}  &
				\begin{tabular}{c} Ours \end{tabular} \\
				
				\midrule
				\multirow{3}{*}{VOT-19}
				& Acc.$\uparrow\ $
				& 0.600 & 0.594
				& 0.580 & {\color{blue}0.603}
				& 0.570 & {\color{green} 0.601 }
				& 0.590
				& {\color{red}0.649}  \\
				
				& Rob.$\downarrow\ $
				& 0.467 & 0.461
				& 0.446 & 0.411
				& {\color{blue}0.366} & 0.386
				& {\color{green}0.376}
				& {\color{red}0.326}  \\
				
				& EAO$\uparrow\ $
				& 0.299 & 0.287
				& 0.292 & 0.292
				& 0.313 & {\color{blue}0.350}
				& {\color{green}0.327}
				& {\color{red}0.356} \\
				
				\bottomrule
		\end{tabular}}%
	\end{center}
	\caption{Results on VOT2019. Top-3 results in each row are colored in red, blue and green, respectively.}
	\label{tab:vot19_results}
\end{table*}

\begin{table*} [t]
	
	\begin{center}
		\tabcolsep=2pt
		\resizebox{2.1\columnwidth}{!}{ %
			\begin{tabular}{c c  c c c c c c c c c c}
				\toprule
				\multicolumn{2}{c}{Trackers} &
				\begin{tabular}{c} CFNet \cite{CFNet} \\  \end{tabular} &
				\begin{tabular}{c} DaSiamRPN \cite{dasiam} \\  \end{tabular} &
				\begin{tabular}{c} SiamRPN++ \\  \end{tabular} &
				\begin{tabular}{c} ATOM \\  \end{tabular} &
				\begin{tabular}{c} DiMP-18 \\ \end{tabular}  &
				\begin{tabular}{c} DiMP-50 \\   \end{tabular}  &
				\begin{tabular}{c} D3S \\ \end{tabular}  &
				\begin{tabular}{c} Ocean-off \\  \end{tabular}  &
				\begin{tabular}{c} Ours \end{tabular} \\
				
				\midrule
				\multirow{2}{*}{GOT-10K}
				
				& SR\textsubscript{.75}$\uparrow\ $
				& 14.4 & 27.0
				& 32.5 & 40.2
				& 44.6  & {\color{blue}49.2}
				& {\color{green}46.2} & -
				& {\color{red}52.2} \\
				
				& AO$\uparrow\ $
				& 37.4 & 48.3
				& 51.8 & 55.6
				& 57.9  & {\color{blue}61.1}
				& {\color{green}59.7} & 59.2
				& {\color{red}61.5}  \\
				
				\bottomrule
		\end{tabular}}%
	\end{center}
	\caption{Results on GOT-10K. Top-3 results in each row are colored in red, blue and green, respectively.}
	\label{tab:got10k_results}
	
\end{table*}

\begin{table*} [!h]
	\tiny
	\begin{center}
		\tabcolsep=2pt
		\resizebox{2.1\columnwidth}{!}{ %
			\begin{tabular}{c c  c c c c c c c c}
				\toprule
				\multicolumn{2}{c}{Trackers} &
				\begin{tabular}{c} UpdateNet \\  \end{tabular} &
				\begin{tabular}{c} SPM \\  \end{tabular} &
				\begin{tabular}{c} SiamRPN++ \\  \end{tabular} &
				\begin{tabular}{c} ATOM \\  \end{tabular} &
				\begin{tabular}{c} DiMP-50 \\ \end{tabular}  &
				\begin{tabular}{c} Retina-MAML \\   \end{tabular}  &
				\begin{tabular}{c} D3S \\ \end{tabular}  &
				\begin{tabular}{c} Ours  \end{tabular} \\
				
				\midrule
				\multirow{2}{*}{TrackingNet}
				
				& Prec.$\uparrow\ $
				& 62.5 & -
				& {\color{blue}69.4} & 64.8
				& {\color{green}68.7} & -
				& 66.4
				& {\color{red}69.7}  \\
				
				& Norm. Prec.$\uparrow\ $
				& 75.2 & -
				& {\color{blue}80.0} & 77.1
				& {\color{red}80.1} & 78.6
				& 76.8
				& {\color{green}79.4}  \\
				
				& Succ.$\uparrow\ $
				& 67.7 & 71.2
				& 73.3 & 70.3
				& 74.0 & 69.8
				& 72.8
				& {\color{red}74.2}  \\
				
				\bottomrule
		\end{tabular}}%
	\end{center}
	\caption{Results on TrackingNet. Top-3 results in each row are colored in red, blue and green, respectively.}
	\label{tab:tk_results}
\end{table*}

\begin{table*} [!h]
	\begin{center}
		\tabcolsep=2pt
		\resizebox{2.1\columnwidth}{!}{ %
			\begin{tabular}{c c  c c c c c c c c c c c}
				\toprule
				\multicolumn{2}{c}{Trackers} &
				\begin{tabular}{c} DPMT\cite{xie2020hierarchical} \\  \end{tabular} &
				\begin{tabular}{c} SuperDiMP \cite{sdimp} \\  \end{tabular} &
				\begin{tabular}{c} DiMP \\  \end{tabular} &
				\begin{tabular}{c} ATOM \\  \end{tabular} &
				\begin{tabular}{c} SiamMargin \cite{Kristan2019a}\\  \end{tabular} &
				\begin{tabular}{c} SiamMask \\  \end{tabular} &
				\begin{tabular}{c} STM \\  \end{tabular} &
				\begin{tabular}{c} DET50 \cite{Kristan2019a}\\ \end{tabular}  &
				\begin{tabular}{c} Ocean\\   \end{tabular}  &
				\begin{tabular}{c} D3S \\ \end{tabular}  &
				\begin{tabular}{c} Ours \end{tabular} \\
				
				\midrule
				\multirow{3}{*}{VOT-20}
				& Mask
				& - & -
				& - & -
				& \checkmark & \checkmark
				& \checkmark & \checkmark
				& \checkmark & \checkmark
				& \checkmark  	\\
				
				& Acc.$\uparrow\ $
				& 0.492 & 0.492
				& 0.457 & 0.462
				& 0.698 & 0.624
				&  {\color{red}0.751} & 0.679
				& 0.693 & {\color{green}0.699}
				& {\color{blue}0.711}  \\
				
				& Rob.$\downarrow\ $
				& 0.745 & 0.745
				& 0.740 & 0.734
				& 0.640 & 0.648
				& 0.574 & {\color{red}0.787}
				& 0.754 & {\color{green}0.769 }
				& {\color{blue}0.776}  \\
				
				& EAO$\uparrow\ $
				& 0.303 & 0.305
				& 0.274 & 0.271
				& 0.356 & 0.321
				& 0.308 & {\color{blue}0.441}
				& 0.430 & {\color{blue}0.439}
				& {\color{red}0.453} \\
				
				\bottomrule
		\end{tabular}}%
	\end{center}
	\caption{Results on VOT2020. ``Mask''
		denotes that prediction format is mask. Top-3 results in each row are colored in red, blue and green, respectively.}
	\label{tab:vot20_results}
\end{table*}

\begin{figure*}[!h]
	\centering{\includegraphics[scale = 0.70]{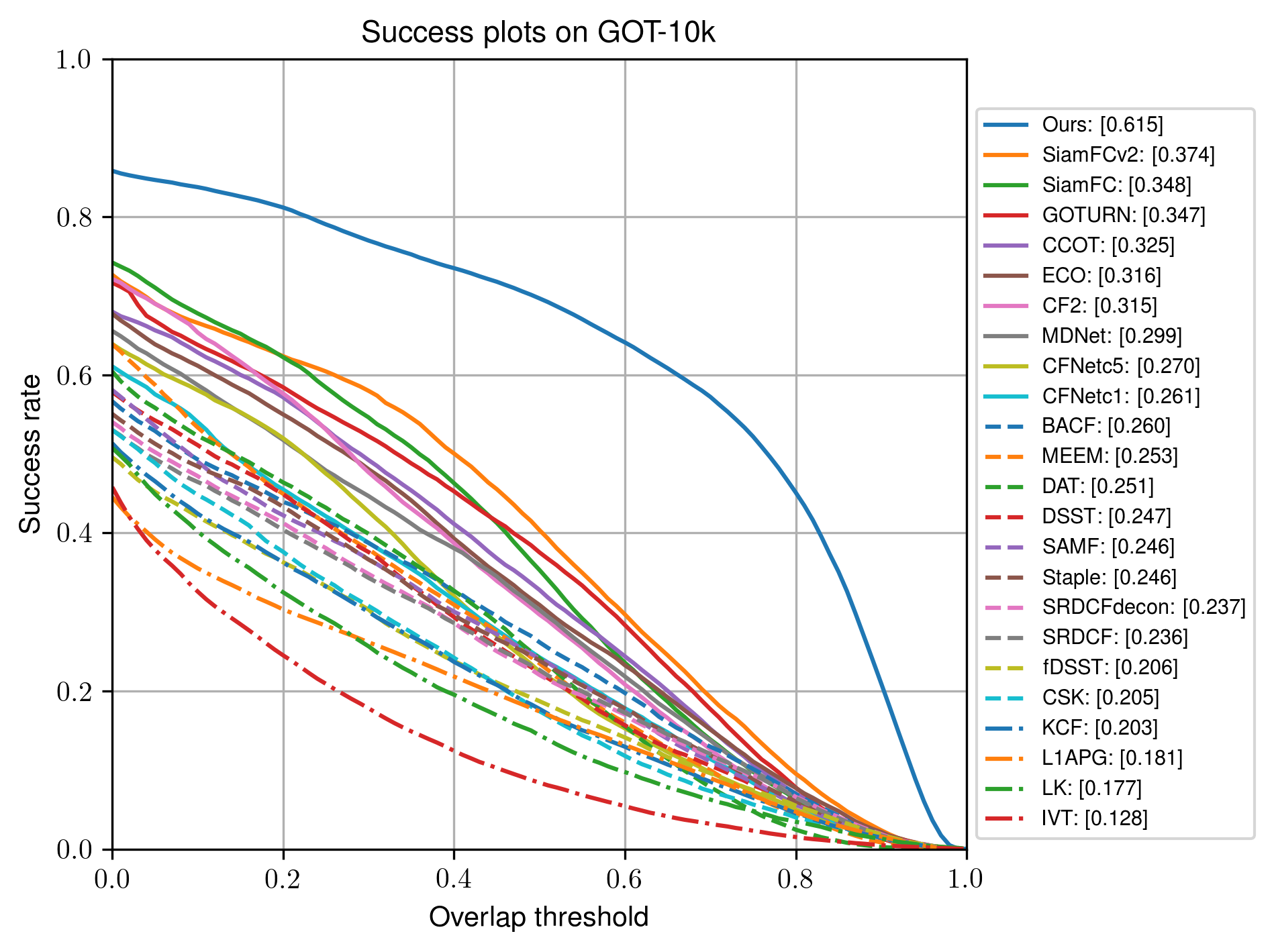}}
	\caption{The success rate plot on GOT-10K that includes a large-scale generic objects in the wild. Results of other trackers are from reports of 25 public entries on GOT-10K official benchmark.
	}
	\label{fig:got}
\end{figure*}

\paragraph{Testing Phase}
During inference, the sampling interval in appearance memory network is set to 5. The output of our model is segmentaion map and will be transferred to rotated box for tracking task.

\paragraph{Box-to-Segmentation Strategy}
Segmentation-based tracker has to be robust towards bounding box initialization. By adding $1$\% bounding box label data called box-mask, we train our model to segment pseudo-mask with only one box-mask stored in memory networks. Then, the pseudo-segmentation mask will replace the box-mask cyclically in memory networks during tracking. This training and testing strategy greatly boosts the tracking performance.

\paragraph{The whole Tracking Process}
The tracker is initialized on the first frame using the bounding-box format ground truth in VOT benchmarks. 
If a ground-truth bounding box is available, the AMN follows the initialization procedure proposed in the box-to-segmentation strategy. 
Our procedure is similar to the procedure proposed in~\cite{lukezic2020d3s}. The geometric center of ground-truth bounding box provides the training samples for SMN initialization process. The bounding box region is firstly segmented by AMN which models the appearance information of targets during initialization.
On the other hand, if a segmentation mask is available, the tracker will use the mask directly. 


\subsection{Evaluation}\label{sec:evaluation}
Our tracker is compared with a variety of SOTA trackers on six major short-term tracking benchmarks including VOT2016~\cite{hadfield2016visual}, VOT2018~\cite{kristan2018sixth}, VOT2019~\cite{Kristan2019a}, GOT-10k~\cite{got}, TrackingNet~\cite{Trackingnet}, VOT2020~\cite{Kristan2020a} and two VOS benchmark datasets DAVIS16~\cite{perazzi2016benchmark} and DAVIS17~\cite{pont20172017}.
We select representative benchmarks based on their prediction output formats and challenging factors, which can be categorized into tracking with rotated/axis-aligned bounding box formate, pixel-wise tracking and VOS.

\subsubsection{Tracking with Rotated Bounding Box Format} The VOT datasets are the most challenging and convincing evaluation benchmark in VOT community.
VOT2016, VOT2018 and VOT2019 are widely-used benchmarks for VOT. Each of them contains 60 sequences with various challenging factors.
The three datasets are annotated with the rotated bounding boxes, and a reset-based methodology is applied for evaluation.
For both benchmarks, trackers are measured in terms of accuracy (A), robustness (R), and expected average overlap (EAO).

\subsubsection{Tracking with Axis-aligned Bounding Box Format}  We also evaluate our tracker in the axis-aligned bounding box annotated visual tracking benchmarks, i.e., GOT-10K~\cite{got} and TrackingNet~\cite{Trackingnet}. Axis-aligned bounding box annotation is widely used among object detection and tracking benchmarks.
GOT-10K is a recently released large-scale dataset (10,000 videos in train subset and 180 in both validate and test subset) with 1.5 million bounding boxes. Average overlap (AO), success rates at 75\% threshold ($SR_{75}$) and 50\% threshold ($SR_{50}$) are the three ranking metrics.
TrackingNet~\cite{Trackingnet} contains 30,000 sequences with 14 million dense annotations and 511 sequences in the test set. It covers diverse object classes and scenes, requiring trackers to have both discriminative and generative capabilities.

\subsubsection{Pixel-wise Tracking}
VOT2020~\cite{Kristan2020a} adopts a new testing protocol compared to 2019 that the target position is encoded by a segmentation mask. Moreover, VOT2020~\cite{Kristan2020a} benchmark introduces a new evaluation methodology for the promising pixel-wise tracking paradigm which requires trackers to robustly track the target while predicting an accurate mask. Segmentation-based trackers need to performs well both in segmentation accuracy and challenging scenarios, e.g., fast motion, distractors and blur. Accuracy (A), robustness (R) and expected average overlap (EAO) are three metrics to evaluate trackers.

\begin{figure}[t]
	\centering{\includegraphics[scale = 0.70]{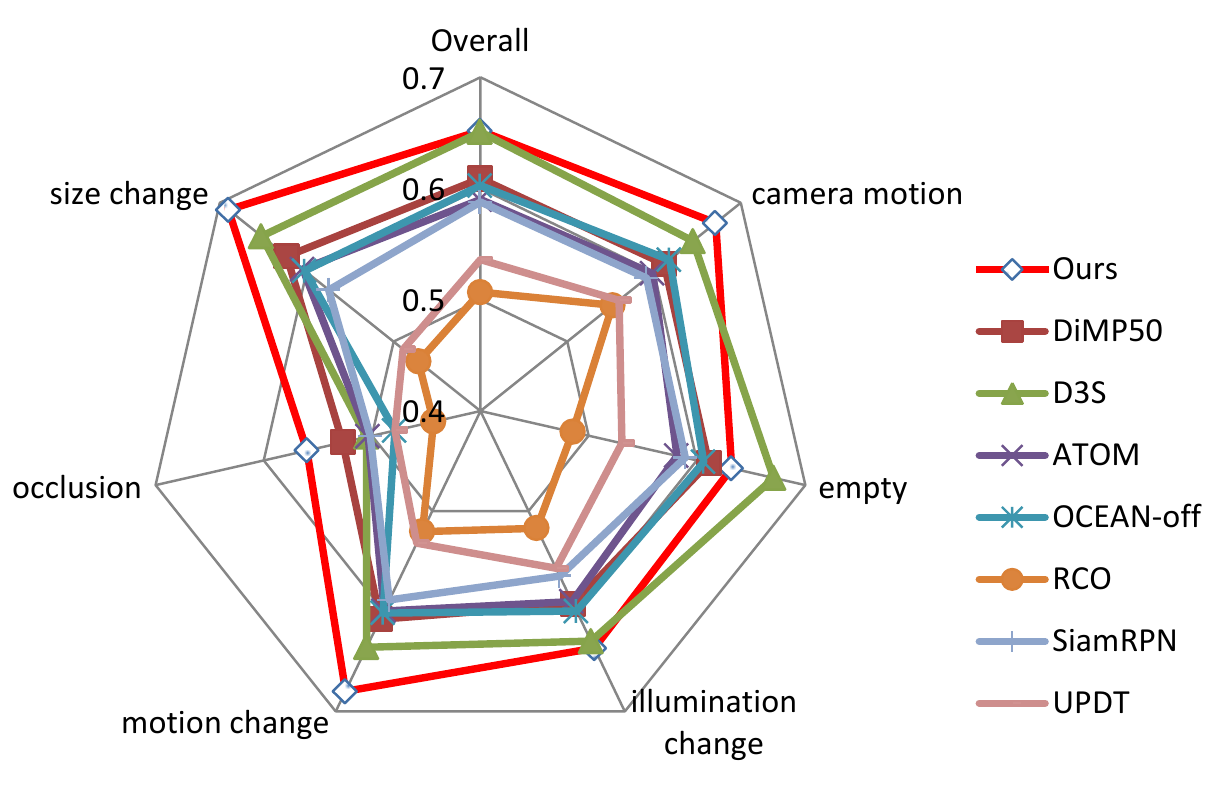}}
	\caption{Comparison of accuracy on VOT2018 for the following visual attributes: camera motion, illumination change, occlusion,	size change and motion change. Empty means frames do not belong to any of the five attributes.		
	}
	\label{fig:vot18}
\end{figure}

\subsubsection{VOS} We also evaluate our tracker in two semi-supervised VOS benchmarks, i.e., DAVIS16\&17~\cite{perazzi2016benchmark}~\cite{DAVIS}, following official test protocol: mean Jaccard index ($J_{M}$) and mean F-measure ($F_{M}$).  $J_{M}$ describes the region similarity while ($F_{M}$) measures the contour accuracy of the predictions.  VOS datasets typically have few challenging factors like distractors and motion blur,  which focus more on segmentation quality.

\textbf{VOT2016.} The VOT2016 top performers CCOT~\cite{CCOT} and TCNN~\cite{TCNN}, two recent SOTA segmentation-based trackers D3S~\cite{lukezic2020d3s} and SiamMask~\cite{siammask}, and most recently published SOTA deep learning-based trackers SiamRPN++~\cite{siamrpn++}, SPM~\cite{spm}, UpdateNet~\cite{updatenet}, and ATOM~\cite{atom} are compared with our trackers.
Fig.~\ref{fig:16eao} and Table \ref{tab:vot16_results} show that our tracker outperforms all trackers on all three measures by a large margin. In terms of EAO, our tracker outperforms the strongest SOTA tracker D3S by 4.2\% and ATOM by 10.2\%. To the best of our knowledge, our tracker is the first one among published papers which breaks through the 0.53 in EAO and 0.68 in accuracy together.

The VOT2016 dataset contains segmentation labels and challenging factors, such as dim and blur targets. 
We have thus compared our tracker with the most recent segmentation tracker D3S in terms of the precision of predicted rotated bounding-box.  
Ours achieves a 0.684 precision, while D3S IoU is 0.66. Promotion from accuracy improvements also raise the robustness performance of our tracker, which achieves 0.121 robustness comparing to that of D3S (0.131 robustness).

\begin{table}[t]
	\setlength\tabcolsep{8pt}
	\small
	\centering
	\caption{Comparison with segmentation-based trackers and VOS methods on DAVIS16 and DAVIS17.}
	\label{tab:davis_compare}
	\setlength{\tabcolsep}{1.6 mm}{
		\scalebox{1.0}{
			\begin{tabular}{l ccc ccc}
				\hline
				& $\mathcal{J}_{\mathcal{M}}^{16}$ &   $\mathcal{F}_{\mathcal{M}}^{16}$
				& $\mathcal{J}\& \mathcal{F}^{16}$
				& $\mathcal{J}_{\mathcal{M}}^{17}$ &  $\mathcal{F}_{\mathcal{M}}^{17}$
				& $\mathcal{J} \& \mathcal{F}^{17}$ \\ \hline
				
				Ours
				& 79.0
				& 75.5
				& 77.3
				& 64.8
				& 67.7
				& 66.3
				\\
				
				D3S~\cite{lukezic2020d3s}
				& 75.4
				& 72.6
				& 74.0
				& 57.8
				& 63.8
				& 60.8
				\\
				
				SiamMask~\cite{siammask}
				& 71.7
				& 67.8
				& 69.8
				& 54.3
				& 58.5
				& 56.4
				\\
				\hline
				OnAVOS~\cite{OnAVOS}
				& 86.1
				& 84.9
				& 85.5
				& 61.6
				& 69.1
				& 65.4
				\\
				STM~\cite{STM}
				& 84.8
				& 88.1
				& 86.4
				& 69.2
				& 74.0
				& 71.6
				\\
				MAST~\cite{MAST}
				& -
				& -
				& -
				& 63.3
				& 67.6
				& 65.5
				\\
				
				FAVOS~\cite{FAVOS}
				& 82.4
				& 79.5
				& 80.9
				& 54.6
				& 61.8
				& 58.2
				\\
				
				VM~\cite{VM}
				& 81.0
				& -
				& -
				& 56.6
				& -
				& -
				\\
				
				OSVOS~\cite{OSVOS}
				& 79.8
				& 80.6
				& 80.2
				& 56.6
				& 63.9
				& 60.3
				\\
				
				PLM~\cite{PLM}
				& 75.5
				& 79.3
				& 77.4
				& -
				& -
				& -
				\\
				
				OSMN~\cite{OSMN}
				&  74.0
				&  72.9
				&  73.5
				&  52.5
				&  57.1
				&  54.8
				\\
				\hline
	\end{tabular}}}
	
\end{table}

\textbf{VOT2018.} VOT2018 is the most widely-used benchmark so far. We compared our tracker with all official results from ~\cite{kristan2018sixth} in Fig.~\ref{fig:18eao}. We also compared our tracker with the most recent SOTA trackers: DCFST~\cite{DCFST}, Ocean~\cite{ocean}, D3S~\cite{lukezic2020d3s}, SiamBAN~\cite{siamban}, DiMP~\cite{dimp}, ATOM~\cite{atom}, SiamRPN++~\cite{siamrpn++} and SiamMask~\cite{siammask}.

Our tracker outperforms the VOT2018 top performers LADCF and SiamRPN++. As shown in Fig.~\ref{fig:18eao}, our tracker outperforms all trackers on all three measures by a large margin. In terms of EAO, our tracker outperforms the SOTA tracker LADCF  by 4.2\% and SiamRPN++ by 10.2\%.

As shown in Table \ref{tab:vot18_results}, Our tracker outperforms all recent SOTA trackers in all three metrics. So far, our tracker is the first one among published papers which breaks through the 0.50 in EAO without redundant modules. Our tracker outperforms the D3S in EAO by 1.7\%, SiamMask in accuracy by 1.0\% and Ocean-off by over 2.4\% in robustness.

As shown in Fig.~\ref{fig:vot18}, our tracker is more accurate than other trackers towards challenging factors like occlusion, size and motion changes. Our tracker ranks first on attributes of occlusion, size change, motion change, camera motion and illumination, and ranks second on attributes of empty. This shows that our tracker is robust towards occlusion, size changes and motion changes in the target while having the ability to handle with camera motion and illumination changes.

\textbf{VOT2019.} The VOT2019 sequences were replaced by 20\% video sequences compared to the VOT2018.
Our tracker is compared to the recent prevailing trackers.
As shown in Fig.~\ref{fig:19eao} and Table \ref{tab:vot19_results}, our tracker surpasses all the competitive trackers in three metrics.
Our tracker outperforms the most recently published Siamese correlation tracker Ocean~\cite{ocean} by 2.9\% in EAO. The accuracy of our tracker outperforms the ATOM by 4.6\%.
The results demonstrate that our tracking architecture has better performance towards both Siamese correlation trackers and filter-based trackers.

\begin{figure}[t]
	\centering{\includegraphics[scale = 0.50]{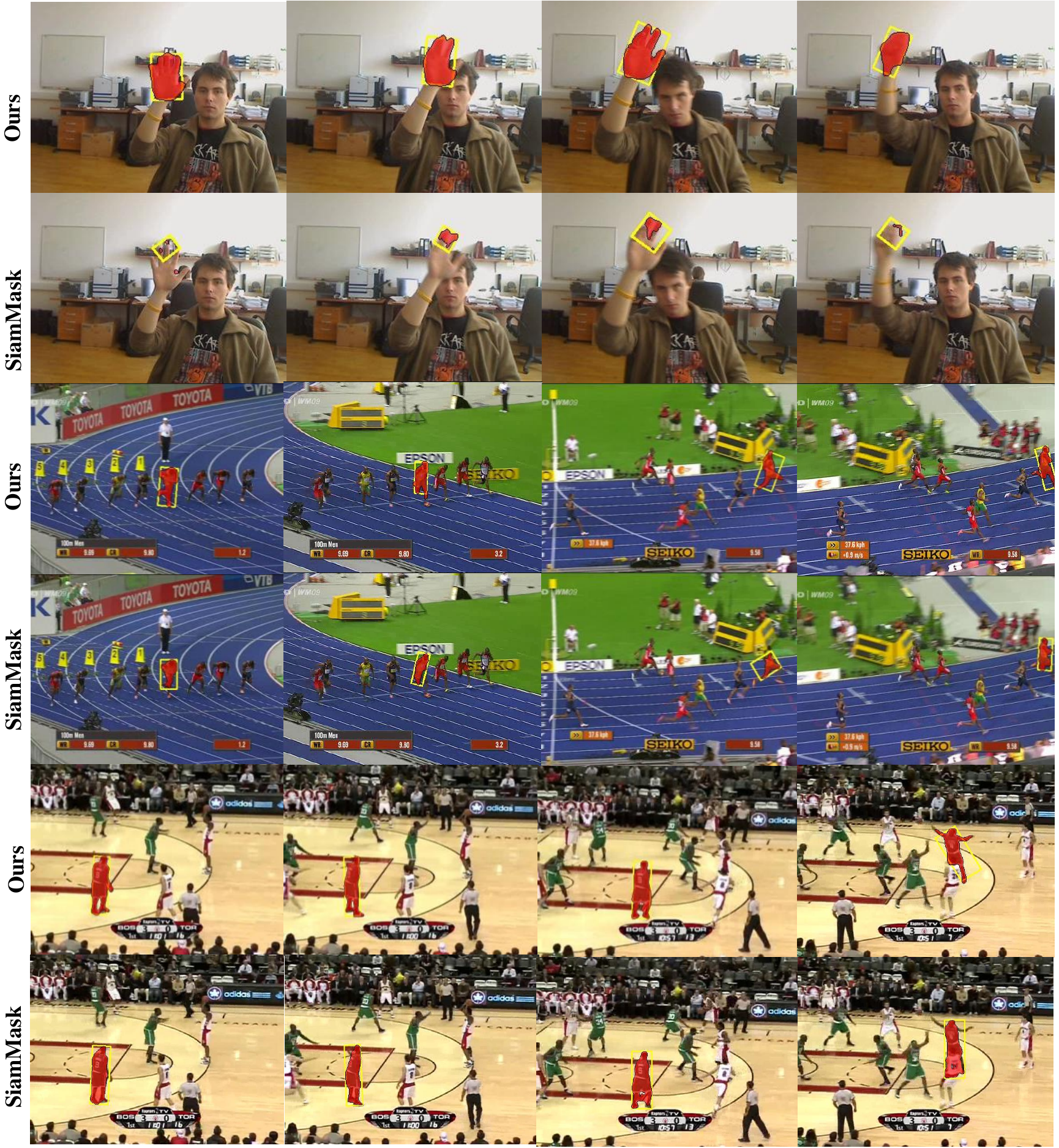}}
	\caption{Ours vs. SiamMask. Our tracker has better generalization ability towards unseen objects such as hand. Moreover, ours is more discriminative towards distractors and achieves a better performance on predicting object contour. }
	\label{fig:siammask}
\end{figure}

\begin{figure*}[t]
	\centering{\includegraphics[scale = 0.59]{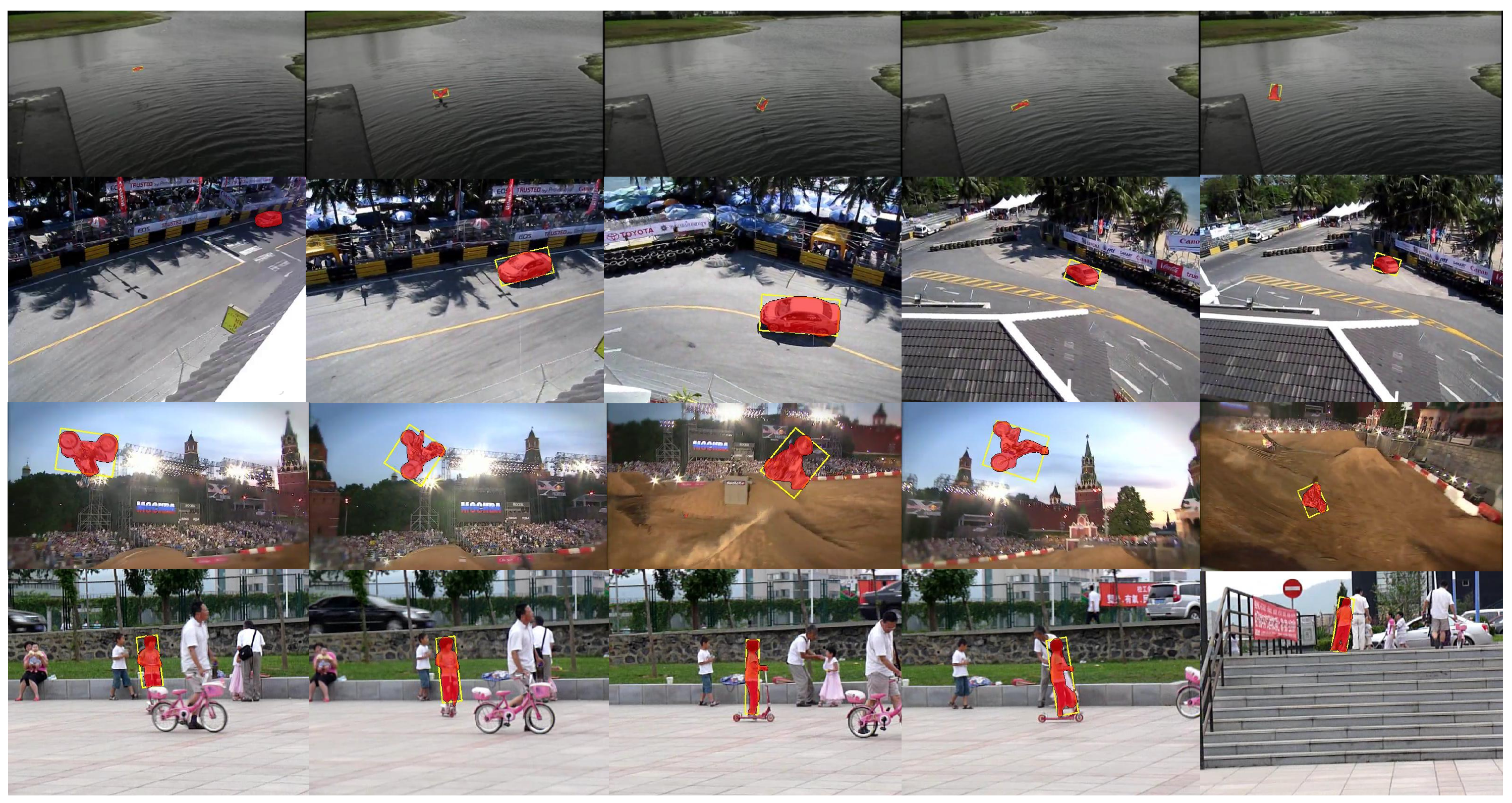}}
	\caption{Qualitative examples of tracking and segmentation. Video sequences are collected from the VOT2018 benchmark. Output of our tracker is segmentation mask in red color. A bounding box is generated from the predicted  mask and shown in yellow. Our tracker can handle with challenging scenarios, such as the dim targets, fast motion and appearance variations}
\end{figure*}

\begin{figure}[t]
	\centering{\includegraphics[scale = 0.38]{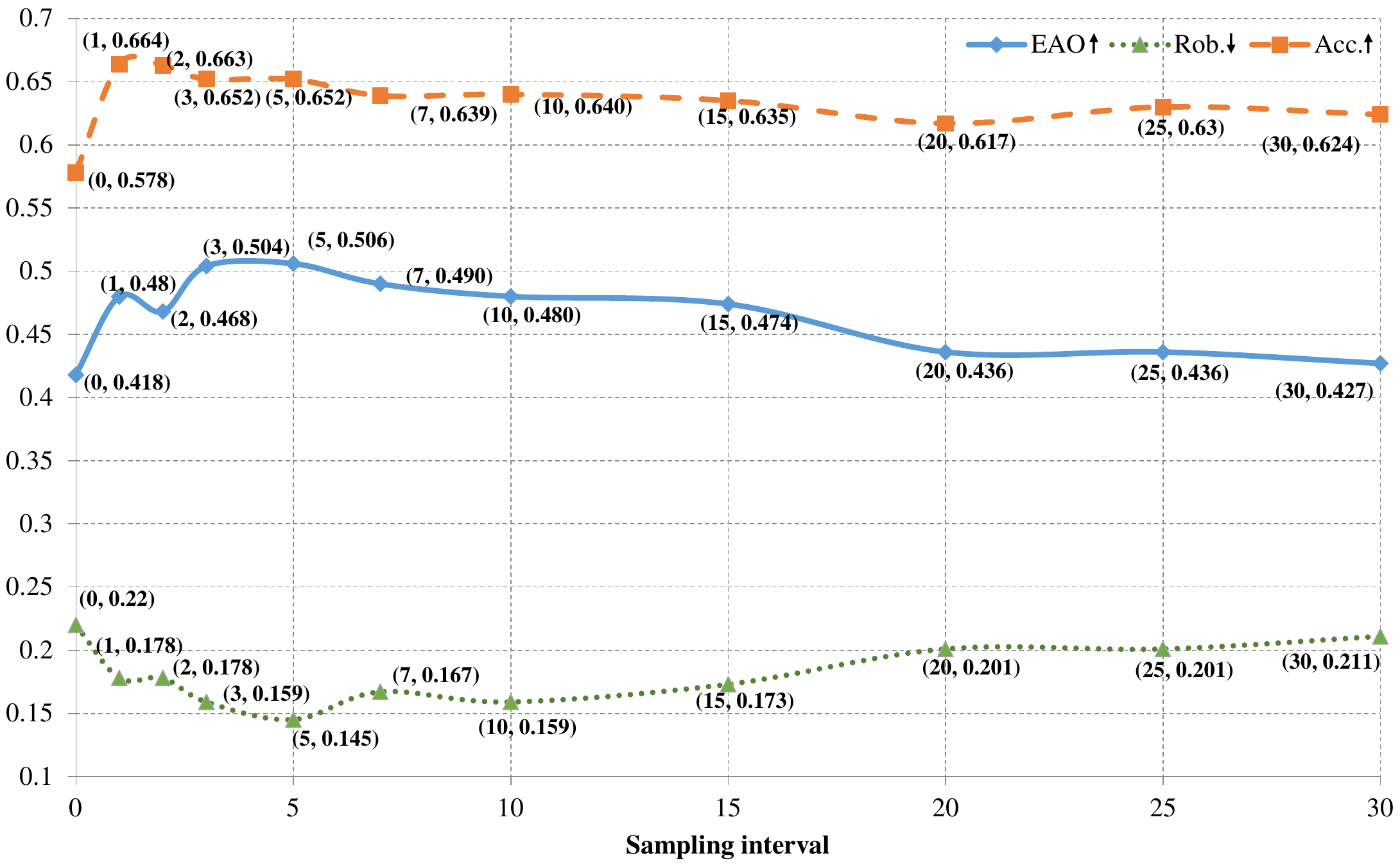}}
	\caption{Time interval indicates the sampling interval of memory network. Zero interval indicates that only the first frame and its ground truth is stored. Up-arrow (down-arrow) indicates higher (lower) is better. }
	\label{fig:interval}
\end{figure}

\begin{table*}[t]
	\normalsize
	\centering
	\caption{Ablation study on VOT2018 and DAVIS16.}
	\setlength{\tabcolsep}{2.5 mm}{
		\begin{tabular}{l|ccccccccc}
			\hline	
			Last Add.  & \checkmark &  & \checkmark  & \checkmark    & \checkmark   & \checkmark  & \checkmark    & \checkmark   \\
			
			Box2Seg  & \checkmark  & \checkmark  &   & \checkmark  & \checkmark  & \checkmark  & \checkmark  & \checkmark   \\
			
			Interv.  & 5 & 5  &5  & 10  & 5  &  5 & 15  &  20  \\
			
			Filter Samp.  & \checkmark  & \checkmark   & \checkmark  & \checkmark   &   & \checkmark   &   &    \\
			
			Pos. Encod.  & sum & sum  & sum & sum  & sum & cat  & sum & sum   \\ \hline

			A $\uparrow\ $
			& 0.652
			& 0.665
			& 0.635
			& 0.640
			
			& 0.627
			& 0.650
			& 0.622
			& 0.620  \\
			
			R $\downarrow\ $
			& 0.145
			& 0.173
			& 0.164
			& 0.159
			
			& 0.210
			& 0.150
			& 0.225
			& 0.227   \\
			
			EAO $\uparrow\ $
			& 0.506
			& 0.467
			& 0.492
			& 0.480
			
			& 0.421
			& 0.486
			& 0.410
			& 0.402   \\ \hline
			
			$\mathcal{J} \& \mathcal{F}^{16}$ $\uparrow\ $
			& 77.3
			& 70.3
			& -
			& 67.8
			
			& 69.1
			& -
			& 66.4
			& 63.6   \\ \hline
			
	\end{tabular}}
	
	\label{tab:vot18ablative}
\end{table*}

\textbf{GOT-10K.} GOT-10K is a recently released large-scale dataset consisting of 10K video segments and 1.5 million classical axis-aligned bounding boxes. As shown in Table \ref{tab:got10k_results}, our tracker outperforms all competing trackers in two metrics and achieves the SOTA $AO$ score of 61.5. Our tracker improves the $SR_{75}$ by 3.0\% over the SOTA filter-based tracker DiMP-50, while outperforming DiMP-50 by 0.4\% in terms of $AO$.
Comparing to the Siamese correlation trackers, our tracker outperforms the Ocean by 2.3\% in terms of $AO$.
Compared to the SOTA segmentation-based tracker D3S, our tracker has improvements of 3.0\% on $AO$ and nearly 13.0\% improvements on $SR_{75}$, demonstrating its ability to tracking objects over complex scenes.

As shown in Fig.~\ref{fig:got}, compared results are from GOT-10K official website. Our tracker outperforms other trackers by a large margin in terms of success rate which fully demonstrates that our segmentation-based tracker also can achieve SOTA performance results in classical axis-aligned bounding box annotation benchmark.

\textbf{TrackingNet.} We further evaluate our tracker on the large-scale TrackingNet. As shown in Table \ref{tab:tk_results}, our tracker outperforms the strongest filter-based tracker DiMP-50 by 0.2\% in AUC while our accuracy surpasses the strongest segmentation-based tracker D3S by 3.3\%.
Compared to the SOTA segmentation-based tracker D3S, our tracker has improvements of 3.4\% on $Norm. Prec.$ and nearly 2.0\% improvements on $Succ.$, demonstrating its considerable improvements among segmentation-based trackers.

\textbf{VOT2020.} Recently, the tracking community starts focusing on replacing the classical rectangle box with a segmentation mask to accurately represent the target. The new evaluation protocol introduced  by VOT2020~\cite{Kristan2020a} is specifically designed for segmentation-based trackers. Our tracker is compared to 6 SOTA trackers with segmentation outputs and 4 trackers with classical bounding box outputs. All results are from  VOT2020 official report~\cite{Kristan2020a} or tested by the released official toolkit.

As listed in Table \ref{tab:vot20_results}, our tracker surpasses all the trackers in terms of EAO measure. Ours outperforms the top SOTA tracker DET50~\cite{Kristan2020a} by 1.2\% (0.453 vs. 0.441). Moreover, ours significantly outperforms the top SOTA VOS method STM~\cite{oh2019video} by 14.5\% in terms of EAO (0.453 vs. 0.308). The VOT2020 benchmark with segmentation mask label is specifically designed for pixel-wise tracking, and our tracker takes dominant position in this benchmark with promising SOTA performance on EAO measure.
It can be observed that the top-ranked VOS method STM performs much worse than the top-performing VOT approaches. The STM performs segmentation without considering tracking which results in less robust performance. Ours considers both robustness tracking and accurate segmentation which can handle with various challenging factors in tracking scenarios. Thus, our tracker outperforms STM in robustness by a large margin (0.776 vs. 0.574).

\textbf{DAVIS16\&17.} The performance is evaluated by two metrics averaged over the sequences following official test protocol: mean Jaccard index ($J_{M}$) and mean F-measure ($F_{M}$).
Our tracker is compared with the SOTA segmentation-based trackers and competitive VOS methods.

From Table~\ref{tab:davis_compare}, we can observe that our tracker outperforms the SOTA segmentation-based trackers D3S and SiamMask~\cite{siammask} by a large margin.
On the more challenging benchmark DAVIS17, our tracker even outperforms all the methods specialized to VOS task except for STM in terms of mean $J\&F$.
Compared to D3S, which also belongs to discriminative segmentation-based tracker, our approach obtains gains of 3.3/5.5\% on J \&F for DAVIS16/17, respectively. Furthermore, we achieve better performance than SiamMask on all criteria of DAVIS16/17, demonstrating the strong segmentation ability of our approach.

\subsection{Ablation Study}\label{sec:ablation}

To further show our contributions, we conduct comprehensive ablation studies on VOT2018 and DAVIS16. The performance on tracking and VOS benchmarks can address the robust tracking and accurate segmentation ability of our tracker, respectively.

\textbf{Temporal Information.} In order to show the effectiveness of utilizing temporal information, we set different sampling interval of the AMN. When sampling interval is 0, our tracker is the same as template-matching methods where only the first frame is used. As shown in Fig.~\ref{fig:interval}, the all three measures drop by a large margin in comparison to the modes utilizing temporal information.
No temporal information used causes 6.2\% performance drop in EAO, 8.6\% drop in Accuracy and 4.2\% drop in Robustness in contrast to storing every sample in memory network. It further validates  the superiority of our tracking architecture to the template-based trackers.

The amount of samples stored in memory network also matters. When  sampling interval is 1, our trackers reaches the top accuracy performance which is 0.663. Performance of EAO reaches the top which is 0.506 when sampling interval is 5 frames. Comparing to the 5 frames interval, 30 frames interval which is sparse reduces the EAO by 7.9\%. When the last frame always be added to the AMN, our tracker boosts its overall performance EAO by 3.9\% and robustness performance by 2.8\% when sampling interval is 5 frames.

\textbf{Box-to-Segmentation.} As shown in Tab.~\ref{tab:vot18ablative}, the box-to-segmentation training and testing strategy improves the EAO value by 1.4\% and the accuracy rises from 0.635 to 0.652. The performance of robustness is improved by 0.9\% (0.173 vs. 0.164).
Box-to-Segmentation strategy aims to equip tracker with external segmentation ability which is suitable for visual tracking problem.
Experiment results show that this strategy mitigates the inaccurate effect of bounding box initialization during tracking.

\textbf{Positional Encoding.} Inspired by CoordConv~\cite{liu2018intriguing}, we concatenate two coordinate channels to read-out features. On the other hand, we simply do positional encoding as that in natural language processing. We add the single spatial  matrix to the read-out features.
As shown in Tab.~\ref{tab:vot18ablative}, adding spatial matrix to the read-out features outperforms the concatenating way by 2\% in terms of EAO. Thus, we choose adding style as our positional encoding way for its simplicity and effectiveness.

\textbf{Sample Filtering.} As shown in Table~\ref{tab:vot18ablative}, the collaboration between the spatio-appearance memory networks is significant to the overall performance. Without the samples filtered from SMN, the EAO drops from 0.506 to 0.421 when sampling interval equals to 5. The mean of $J\&F$ on DAVIS16 also reduces from 77.3 to 69.1. This indicates that one single memory network cannot handle these challenging tracking scenarios separately. Ours can handle with both VOT and VOS tasks while keeping fast inference speed

\section{Conclusion}
In this paper, we have proposed a novel segmentation-based tracking architecture which can capture the rich temporal information by learning an effective spatio-appearance network. To be specific, we have designed an appearance memory network and a spatial memory network, which mutually promote to significantly boost the overall tracking performance. Finally, we have leveraged the box-to-segmentation strategy to reduce the gap between VOT and VOS, further boosting the segmentation accuracy.
%
%
%
Without bells and whistles, our tracker has achieved state-of-the-art performance on six large-scale challenging tracking benchmarks with different prediction formats, especially in VOT2020 benchmarks which is designed for segmentation-based trackers.
In the future, we will further improve the spatio-appearance memory network architecture, especially in the aspect of efficient memory management and make the two memory networks more collaborative and unified. We hope to develop a model that can achieve SOTA performance on both VOT and VOS tasks while keeping real-time inference speed.


\ifCLASSOPTIONcaptionsoff
  \newpage
\fi

\bibliographystyle{IEEEtran}
\bibliography{ref}

\end{document}